\documentclass[conference]{IEEEtran}
\usepackage{microtype}
\usepackage{graphicx}
\usepackage{booktabs} % for professional tables
\usepackage{placeins} 
\usepackage{hyperref}

\usepackage{array}
\usepackage[caption=false,font=normalsize,labelfont=sf,textfont=sf]{subfig}
\usepackage{textcomp}
\usepackage{stfloats}
\usepackage{url}
\usepackage{verbatim}
\usepackage{graphicx}
\usepackage{makecell}
\usepackage{amsthm}
\usepackage{hyperref}
\usepackage{mathrsfs}
\usepackage{enumitem}
\usepackage{booktabs}

\usepackage{xcolor}
\usepackage{multirow}
\usepackage{boldline} 
\usepackage{algorithm}  
\usepackage{algorithmic}
\usepackage{bm} 

\usepackage{tikz}
\usepackage{cancel}
\usepackage{multirow}

\usepackage{array}
\newcolumntype{I}{!{\vrule width 1pt}}
\newcommand*{\Hline}[0]{%
\noalign{\global\setlength{\arrayrulewidth}{1.5pt}}%
\hline
\noalign{\global\setlength{\arrayrulewidth}{0.4pt}}%
}

\usepackage{amsmath,amssymb,amsfonts}

\theoremstyle{plain}

\theoremstyle{definition}

\theoremstyle{remark}

\begin{document}

\title{Spatio-temporal Multivariate Time Series Forecast with Chosen Variables}

\author{
\IEEEauthorblockN{Zibo Liu}
\IEEEauthorblockA{\textit{CISE} \\
\textit{University of Florida} \\
Gainesville, FL, USA \\
ziboliu@ufl.edu}
\and
\IEEEauthorblockN{ Zhe Jiang}
\IEEEauthorblockA{\textit{CISE} \\
\textit{University of Florida} \\
Gainesville, FL, USA \\
zhe.jiang@ufl.edu}
\and
\IEEEauthorblockN{ Zelin Xu}
\IEEEauthorblockA{\textit{CISE} \\
\textit{University of Florida} \\
Gainesville, FL, USA \\
zelin.xu@ufl.edu}
\and
\IEEEauthorblockN{Tingsong Xiao}
\IEEEauthorblockA{\textit{CISE} \\
\textit{University of Florida} \\
Gainesville, FL, USA \\
xiaotingsong@ufl.edu}
\and
\IEEEauthorblockN{Yupu Zhang}
\IEEEauthorblockA{\textit{CISE} \\
\textit{University of Florida} \\
Gainesville, FL, USA \\
y.zhang1@ufl.edu}
\and
\IEEEauthorblockN{Zhengkun Xiao}
\IEEEauthorblockA{\textit{CISE} \\
\textit{University of Florida} \\
Gainesville, FL, USA \\
xiaoz@ufl.edu}
\and
\IEEEauthorblockN{Haibo Wang}
\IEEEauthorblockA{\textit{Department of Computer Science} \\
\textit{University of Kentucky} \\
Lexington, KY, USA \\
haibo@ieee.org}
\and
\IEEEauthorblockN{Shigang Chen}
\IEEEauthorblockA{\textit{CISE} \\
\textit{University of Florida} \\
Gainesville, FL, USA \\
sgchen@ufl.edu}
}

\maketitle

\begin{abstract}
Spatio-Temporal Multivariate time series Forecast (STMF) uses the time series of
$n$ spatially distributed variables in a period of recent past to forecast their values in a period of near future. It has important applications in spatio-temporal sensing forecast such as road traffic prediction and air pollution prediction. Recent papers have addressed a practical problem of missing variables in the model input, which arises in the sensing applications where the number $m$ of sensors is far less than the number $n$ of locations to be monitored, due to budget constraints. We observe that the state of the art assumes that the $m$ variables (i.e., locations with sensors) in the model input are pre-determined and the important problem of how to choose the $m$ variables in the input has never been studied. This paper fills the gap by studying a new problem of STMF with chosen variables, which optimally selects $m$-out-of-$n$ variables for the model input in order to maximize the forecast accuracy. We propose a unified framework that jointly performs variable selection and model optimization for both forecast accuracy and model efficiency. It consists of three novel technical components: (1) \emph{masked variable-parameter pruning}, which progressively prunes less informative variables and attention parameters through quantile-based masking; (2) \emph{prioritized variable-parameter replay}, which replays low-loss past samples to preserve learned knowledge for model stability; (3) \emph{dynamic extrapolation mechanism}, which propagates information from variables selected for the input to all other variables via learnable spatial embeddings and adjacency information. Experiments on five real-world datasets show that our work significantly outperforms the state-of-the-art baselines in both accuracy and efficiency, demonstrating the effectiveness of joint variable selection and model optimization. 
% Our source code is available at: \url{https://anonymous.4open.science/r/NLT-5E56/}
\end{abstract}

\begin{IEEEkeywords}
Multivariate time series, spatio-temporal forecast, machine learning, budget constraints
\end{IEEEkeywords}

\section{Introduction}\label{sec:introduction}
% \input{tables_figs_tex/partial sensing}
% Partial sensing long-term spatio-temporal traffic forecasting represents a cutting-edge approach in intelligent transportation systems, where the challenge lies in making accurate predictions with minimal data. 

\begin{figure}[htbp]
\centering
\includegraphics[width=0.90\columnwidth]{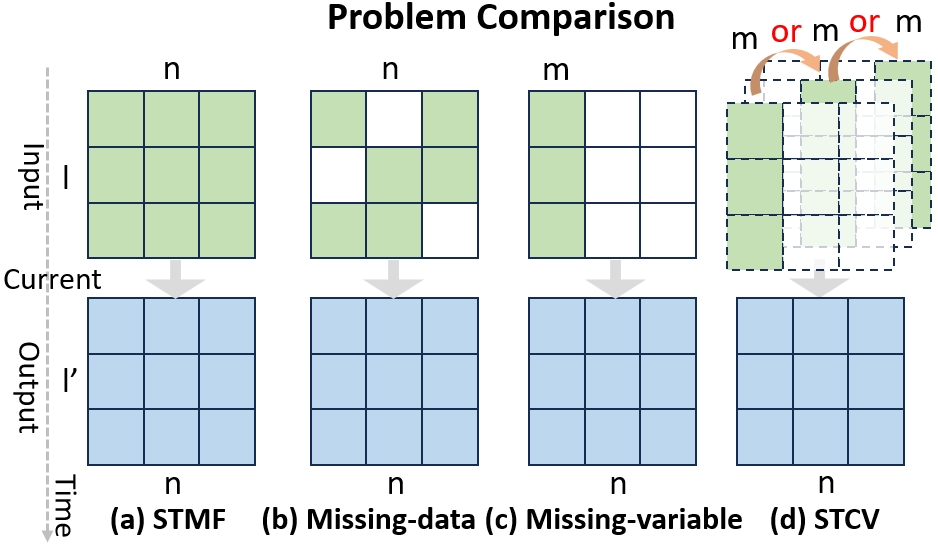}
% \captionsetup{font=small}
\centering
\caption{Problem comparison: (a) STMF, using $n \times l$ input of the past to forecast $n \times l'$ output of the future, (b) STMF with missing data, where some values in the input are missing, (c) STMF with missing variables, where some variables are missing from the input due to budget constraint, using $m \times l$ input to forecast $n \times l'$ output, (d) STCV, allowing optimal selection of the $m$ variables in the input to maximize forecast accuracy. 
%(e) visualization of flow rates of three locations for two days from the PEMS08 dataset.
%The rates are measured as the number of vehicles passing a location in each 5-min time interval. Location 121 had a relatively high flow rate compared to locations 44 and 33. Locations 121 and 33 exhibit regular traffic patterns, with higher flow rates during the day and lower at night. In contrast, location 44 exhibits irregularity from 08/19 15:00 to 20:00. 
} 
\label{fig:fluctuations}
\vspace{-3mm}
\end{figure}
\textbf{New Problem: } 
The problem of {\it Spatio-Temporal Multivariate time series Forecast} (STMF) is to use the time series of $n$ spatially distributed variables in a period $l$ of recent past to forecast their values in a period $l'$ of near future. It has been extensively studied \cite{liu2023spatio,lee2024testam,li2020spatial,Jiang_Wang_Yong_Jeph_Chen_Kobayashi_Song_Fukushima_Suzumura_2023}, which learn their forecast models from the training data of the $n$ time series recorded for adequately long time. Recent studies have explored two variants of STMF. The first one is {\it STMF with missing data} \cite{chen2023biased,cini2021filling,zhang2023trid,tang2020joint}, which deals with sporadically missing values in the $n \times l$ input. The second variant is {\it STMF with missing variables} \cite{10.1145/3580305.3599357,su2024spatial,yu2024ginar}, which deals with permanently missing a subset of variables in the input --- the model input dimensions become $m \times l$ with $n-m$ variables missing, where $m < n$. These variants are motivated from practical applications, and they assume that while the model input for forecasting may miss variables, the training data --- serving as the ground truth for modeling --- have all $n$ time series, which will be justified shortly. The difference between STMF and its variants is illustrated in Figure~\ref{fig:fluctuations}(a)--(c). 

The above problem of STMF with missing variables assumes that the $m$ variables in the model input are given, without considering how to select them from the $n$ variables. This paper studies a new, complementary problem, {\it STMF with Chosen Variables}, denoted as STCV, that addresses the issue of optimally choosing the $m$ variables in the input to maximize the model forecast accuracy. Its difference from the prior studies is illustrated in Figure~\ref{fig:fluctuations}(d). Its practical motivation is given below.

\textbf{Application Motivation: } 
Consider a city with $n$ locations of interest for traffic forecast, which may help monitor potential congestion and provide data for optimizing traffic signing. In this STMF application, sensors can be deployed at the $n$ locations (e.g., street intersections or highway exits) to record traffic volumes in 5-min intervals. A model is learned from the training data, i.e., $n$ time series of sensor data over a long period of recording, e.g., 6 months. When the model is deployed, at each time interval, it may take one past hour of sensor data as input (size $n \times 12$) to forecast one future hour of traffic as output (size $n \times 12$), where one hour has 12 intervals of 5 minutes. 

The case of missing data mirrors sporadic sensor failures. The case of missing variables (i.e., missing sensors) is due to cost saving. The equipment and installation cost of permanent traffic sensors is in thousands of dollars per intersection \cite{retailsensing_vehicle_detection,chatgpt_search_vehicle_detection_cost,google_search_vehicle_detection_cost}, for example, \$10,000–\$30,000 for an inductive loop and \$20,000–\$40,000 for a video-based system, whereas cheap, temporary sensor installation only costs  hundreds of dollars a piece \cite{chatgpt_cheap_search_vehicle_detection_cost}. Considering hundreds or thousands of street intersections in a city, if the budget is limited, one may consider deploying expensive permanent sensors at only a subset of $m$ intersections to save cost, while using cheap, temporary sensors to collect the training data at all locations ---- these temporary sensors will be removed and reused elsewhere after model training, while $m$ permanent sensors will be deployed to provide ongoing input to the model for forecasting. The cost saving can be substantial if one can drive $m$ down to a small fraction of $n$ such as 10\% as our evaluation will achieve. 
%moreover, these temporary sensors can be reused elsewhere or at other times to amortize the cost of data collection. 
The existing work on STMF with missing variables \cite{10.1145/3580305.3599357,su2024spatial,yu2024ginar} assumes the locations of $m$ permanent sensors are pre-determined and investigates how to build a forecast model to forecast traffic at the $n-m$ locations without permanent sensors. This paper is complementary to the existing work by choosing the $m$ locations for optimal deployment of permanent sensors.

Another application that is also used in our evaluation is air quality monitoring, where permanent sensors are deployed in some chosen locations to provide input for a model to forecast the air quality at numerous other locations without permanent sensors. 

\begin{figure}[htbp]
\centering
\includegraphics[width=0.90\columnwidth]{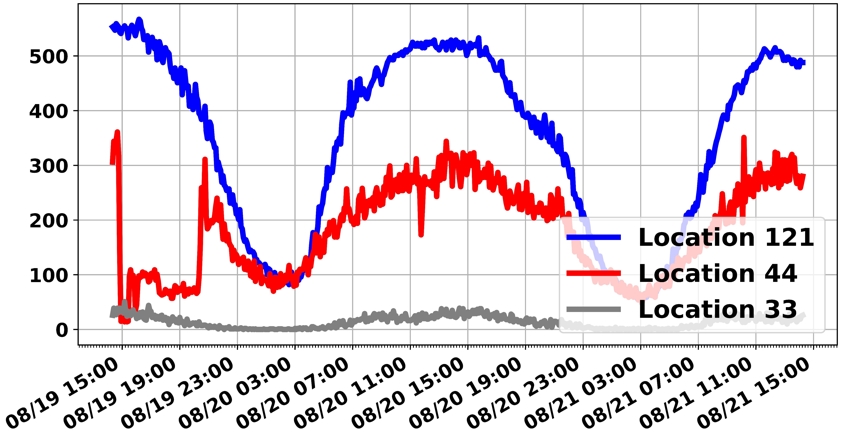}
% \captionsetup{font=small}
\centering
\caption{visualization of traffic volumes at three locations for two days from the PEMS08 dataset.
%(e) visualization of flow rates of three locations for two days from the PEMS08 dataset.
%The rates are measured as the number of vehicles passing a location in each 5-min time interval. Location 121 had a relatively high flow rate compared to locations 44 and 33. Locations 121 and 33 exhibit regular traffic patterns, with higher flow rates during the day and lower at night. In contrast, location 44 exhibits irregularity from 08/19 15:00 to 20:00. 
} 
\label{fig:fig_2}
\vspace{-3mm}
\end{figure}
\textbf{Challenges: } The new problem of STMF with chosen variables (STCV) presents several fundamental challenges. First, the underlying distributions of the variables can vary significantly. As shown in Fig.~\ref{fig:fig_2}, traffic patterns at locations 121, 44, and 33 from the PEMS08 dataset differ notably, which is also observed for numerous other locations, making it difficult to choose where sensors should be replaced so that readings from the chosen locations can accurately forecast other locations. Second, the space of possible variable selections is combinatorially large. Given $n$ variables and a budget of $m$ in the model input, there are $\binom{n}{m}$ possible selections, which makes exhaustive search infeasible, especially for very large scale deployments. 
Third, the data scale can be very large. Consider the traffic forecast application. The model input/output will include a spatio-temporal traffic matrix of tens to hundreds intersections (town), to thousands (city), to tens of thousands (e.g., over 50,000 intersections in NYC) in the space dimension and numerous time intervals in the time dimension, for potentially millions variables, together with additional topological input. This is time series forecast with a very large number of variables. It poses two challenges. One is to accurately forecast the future values of a large number of variables from the past values of a relatively small number of variables. The other is to do so efficiently, both in memory and in inference time, in order to scale to large problem size. 
Finally, as variable selection and model structure evolve during training, the model faces the risk of \emph{catastrophic forgetting}, where previously learned patterns may be lost as variables are removed from the model input. Ensuring stable learning during variable selection presents an additional challenge.

\textbf{Contributions:} 
To address the STCV problem, we propose a new model-learning framework called \emph{Variable-Parameter Iterative Pruning (VIP)}, which consists of three components: masked variable-parameter pruning, dynamic extrapolation, and prioritized variable-parameter replay. The main contributions of this paper are as follows:
%\begin{itemize}
%    \item 

$\bullet\ $\textbf{New problem Introduction.} We introduce and formally define the problem of spatio-temporal multivariate time series forecast with chosen variables, which integrates variable selection and spatio-temporal time series forecast into a unified optimization task under a given budget of variables in the input. This problem setting reflects deployment constraints in real-world applications and has not been systematically studied before.

$\bullet\ $\textbf{Masked variable-parameter pruning}. We design a unified masked pruning mechanism that simultaneously selects variables and prunes redundant attention parameters. This design enables the model to learn compact yet accurate representations, resulting in both improved forecast accuracy and reduced model complexity. Our variable-parameter pruning technique can not only solve the problem of optimal variable selection for forecast accuracy, but also address scalability of large problem size. The model's memory footprint is drastically reduced with variable-parameter pruning and its inference time is also reduced with a smaller model.

%\item 
$\bullet\ $\textbf{Dynamic extrapolation.} We introduce a dynamic extrapolation module that leverages graph structure and node embeddings (that are iteratively learned) to infer values of missing variables, i.e., those that are not selected for the input. This enables accurate forecasting even with highly sparse variable selection.

%    \item 
$\bullet\ $ \textbf{Prioritized variable-parameter replay.} We propose a prioritized replay strategy that buffers and reuses informative samples during training. This mitigates catastrophic forgetting during variable-parameter pruning, ensuring stable optimization and model convergence.

\section{Preliminaries} \label{sec:pre}
\subsection{Problem Definition}
\noindent{\itshape \textbf{Definition 1:}}
Consider a set $N$ of $n$ time series variables. For the road traffic application in the introduction, each variable represents a location of interest, and its time series consists of the traffic volumes at the location in successive time intervals; for the air quality application, each variable represents a location in a city, and its time series may consist of the PM2.5 measurements after successive time intervals. For simplicity, we normalize each time interval as one unit of time. Let $M$ be a subset of variables, i.e., $M \subseteq N$, where $m = |M|$. Let $T$ be a series of time intervals --- for example, $T$ may be $\{t-l+1, ..., t-1, t\}$ of $l$ intervals, where $t$ is the current time. The {\it value matrix} over variables $M$ and time $T$ is defined as $\mathcal{X}_{M, T}= [\mathcal{X}_{i,j}, i \in M, j \in T ]$, where $\mathcal{X}_{i,j}$ is the value of variable $i$ at time $j$. Further denote the time series of variable $i$ over $T$ as $\mathcal{X}_{i, T} = [\mathcal{X}_{i,j}, j \in T]$. Hence, $\mathcal{X}_{M, T}=[\mathcal{X}_{i, T}, i \in M ]$, consisting of $m$ time series for $m$ variables over time $T$. Let $A=[A_{i,j},i,j \in N] \in \mathbb{R}^{n \times n} $ be a {\it spatial adjacency matrix} of the variables in $N$. It is needed for spatio-temperal forecast, where the distributed variables are spatially connected --- for example, the intersections in the road traffic application are interconnected by a road system. The adjacency matrix $A$ defines a graph where each node represents a variable. The $i$th row (column) in $A$ represents variable $i$ (or node $i$). $A_{i, j} \neq 0$ indicates there is a link between node $i$ and node $j$, and its value represents a static connectivity feature of the link. For example, $A_{i,j}=1$ (or 0) may indicate that there is (or not) a road connecting location $i$ and location $j$ \cite{song2020spatial}, and its value may be set proportional to the road distance. In the air quality application, $A$ can be a learnable adjacency matrix \cite{yu2024ginar}.
In summary, $A$ represents the static feature of the multivariate time series system, whereas $\mathcal{X}_{N, T}$ represents the dynamic feature of the system. 

\noindent{\itshape \textbf{Definition 2:} Problem of Spatio-Temporal Multivariate Time Series Forecast with Chosen Variables (STCV).}\label{partial definition} 
The problem of STCV is to build a model that selects a subset $M$ of $m$ variables as the model input and uses their value matrix $\mathcal{X}_{M, T}$ of the recent past to forecast the value matrix $\mathcal{X}_{N, T'}$ in the near future for all variables, where $n > m$, $T' = \{t + 1, ..., t + l'\}$, $T = \{t-l+1, ..., t-1, t\}$, $l'$ is the length of the recent past in the input, $l$ is the length of the near future in the output, and time $t$ is the current time. $M$ is the subset of $m$ chosen variables in the model input, and $M' = N-M$ is the subset of $m'$ remaining variables missing from the input. 

There are two special cases: If $m = n$, the problem becomes the traditional STMF, which was extensively studied \cite{jiang2023spatio,song2020spatial,liu2023spatio,lee2024testam}. If $m < n$ and $M$ is given, it is STMF with missing variables, which was studied recently \cite{yu2024ginar,10.1145/3580305.3599357}. 

Similar to  \cite{yu2024ginar,10.1145/3580305.3599357}, we assume that the training data contains the time series of all variables in both $M$ and $M'$ as the ground truth for model training, which has been justified in the introduction.

\subsection{Temporal Embeddings and Node Embeddings} \label{sec:embeddings}
Learnable temporal embeddings, denoted as $E_N^{temp}$, are used to capture temporal patterns. They are application-dependent. For example, for the road traffic application, they may include time-of-day (tod) embeddings and day-of-week (dow) embeddings \cite{shao2022spatial,liu2023spatio}: Discretize a day into $D = 288$ time steps of 5-minute each, and a week into $W = 7$ days. Each time step in $[0, D)$ is mapped to a trainable time-of-day embedding vector in $\mathbb{R}^{d_{tod}}$, and each day in $[0, W)$ is mapped to a trainable day-of-week embedding vector in $\mathbb{R}^{d_{dow}}$. Given the value  matrix $\mathcal{X}_{N, T} \in \mathbb{R}^{n \times l}$, where $n$ is the number of variables and $l$ is the length of time $T$, we assign temporal embeddings to each entry $\mathcal{X}_{i, j}$. This yields a time-of-day embedding matrix $E^{tod}_N \in \mathbb{R}^{n \times l \times d_{tod}}$ and a day-of-week embedding matrix $E^{dow}_N \in \mathbb{R}^{n \times l \times d_{dow}}$.
%, where each $(i, j)$ pair corresponds to the time-of-day and day-of-week embeddings based on its timestamp. 
We aggregate the above learnable temporal embeddings as $E_N^{temp} =E_N^{tod}||E_N^{dow}$.
In addition, each node $v$ in the graph has a learnable node embedding vector in $\mathbb{R}^{d_v}$; the node embedding matrix is denoted as $E^{node}_N \in \mathbb{R}^{n \times l \times d_v}$, which repeats the node embeddings across time $T$ of length $l$.

\section{Framework of Variable-Parameter Iterative Pruning}
\subsection{Pre-training A Base STMF Model} \label{sec:STMF}
We begin with a base STMF model to address the spatio-temporal multivariate time series forecast problem. The model is denoted as $F(\mathcal{X}_{N,T}, A, \Theta)$, which takes a past value matrix $\mathcal{X}_{N,T}$ from all variables $N$ together with $A$ as input to forecast a future value matrix $\mathcal{X}_{N,T'}$ as output, with the learned model parameters $\Theta$. The model begins with a Multi-Layer Perceptron (MLP) that processes the input data $\mathcal{X}_{N, T}$ and produce a feature embedding $E_N^f \in \mathbb{R}^{n  \times l \times d} $. It then concatenates this with the learnable temporal embeddings $E_N^{temp}$ and the learnable node embeddings $E_N^{node}$ to form aggregated features $ E_N^{all} \in \mathbb{R}^{n  \times l \times q}$, where $q$ is the number of dimensions in the attention layers. $E_N^{all}$ is then passed through a number of temporal attention layers, each denoted as $ATT_t$, and a number of spatial attention layers, each denoted as $ATT_s$, to produce a high-dimensional representation $H_{N,T} \in \mathbb{R}^{n  \times l \times q}$. 
\begin{align}
    E_N^f&=MLP(\mathcal{X}_{N, T})\label{eq:feature emb}\\ 
    E_N^{all}&=E_N^f||E_N^{node}||E_N^{temp}\label{eq:cat emb}\\
    H_{N,T}&=ATT_s(...ATT_s(ATT_t(...ATT_t(E_N^{all})...))...) \label{eq:all att}
\end{align}
where $ATT_t$ and $ATT_s$ are designed to capture the spatio-temporal correlations. Let $H_{N,T}^t \in \mathbb{R}^{n \times l \times q}$ be the output of the first temporal attention, $ATT_t(E_N^{all})$. It is calculated as follows: We obtain query $Q^t$, key $K^t$ and value $V^t$ matrices through the temporal attention layer, where $W_Q^t$, $W_K^t$, $W_V^t \in \mathbb{R}^{q \times q}$ are their learnable parameters, respectively:  
\begin{align}
    Q^t&=E_N^{all}W_Q^t, \,\,K^t=E_N^{all}W_K^t, \,\,V^t=E_N^{all}W_V^t \label{eq:qkv} \\
    H_{N,T}^t&=Softmax\left(\frac{Q^t(K^t)^\top}{\sqrt{d_t}}\right)V^t \label{eq:score}
\end{align}
where $d_k$ is the number of heads in the multi-head mechanism of the attention. We repeat for additional temporal attention layers, each with its input being the output of the previous layer, using that input to substitute $E_N^{all}$ in Eq. \eqref{eq:qkv}-\eqref{eq:score}.   
Let $H_{N,T}^s \in \mathbb{R}^{n \times l \times q}$ be the output of the spatial attention, $ATT_s((H_{N,T}^t)^\top)$, where $\top$ is for transpose. Each spatial attention is implemented similarly, with $ATT_s(H) = ATT_t(H^\top))$ for any input $H$. Let $H_{N,T}^s \in \mathbb{R}^{n \times l \times q}$ be the output after all spatial attentions. Finally, we use an MLP to map  $\hat{H}_{N,T}^s \in \mathbb{R}^{n \times l \times q}$ to the forecast output $\hat{\mathcal{X}}_{N,T'} \in \mathbb{R}^{n \times l'}$:
\begin{align}
     \hat{\mathcal{X}}_{N,T}=MLP(H_{N,T}^s)
     \label{eq:final fc}
\end{align}

\subsection{Variable-Parameter Iterative Pruning}

\begin{figure*}[htbp]
\centering
\includegraphics[width=\textwidth]{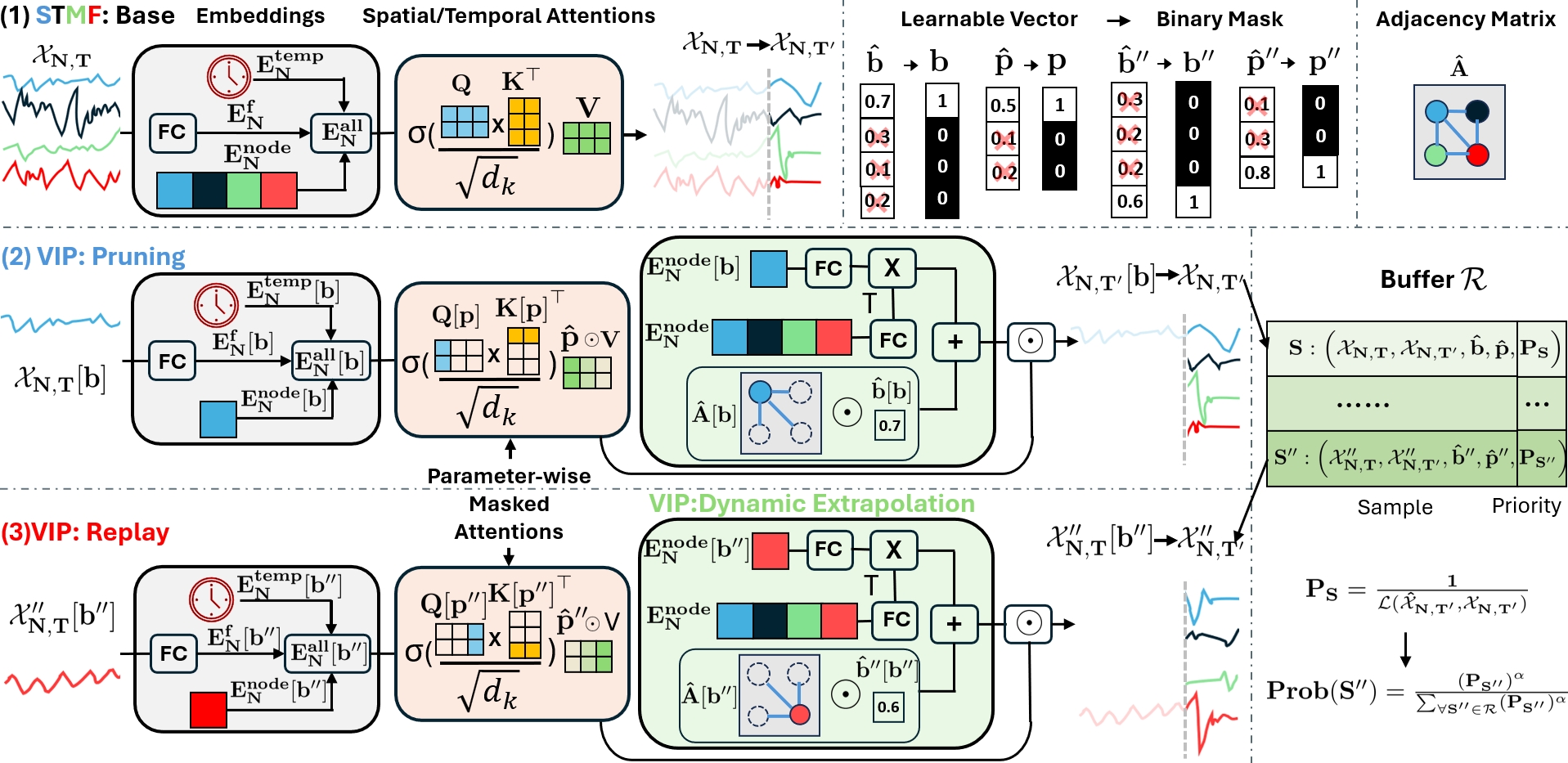}
% \captionsetup{font=small}
\centering
% \caption{Unified spatial-temporal sparsification (USS) contains three components: (1) storing and training the current samples with the loss $\mathcal{L}_{main}$, denoted in blue, (2) replaying and training the previous samples from buffer with the loss $\mathcal{L}_{replay}$, denoted in pink, and (3) increasing the searching space of node/parameter selection through the trainable node vector $\hat{b}$ and the trainable parameter vector $\hat{p}$, from which the node mask $b$ and the parameter mask $p$ are computed, denoted in green.} 

\caption{(1) the base STMF model, shown at the top left of the figure.  
(2) VIP: pruning, shown at the middle left. It learns from the current sample, 
$S: \left(\mathcal{X}_{N,T}, \mathcal{X}_{N,T'}, \hat{b}, \hat{p}, P_{S}\right)$, and then stores the sample in a buffer, shown at the middle right.  
(3) VIP: replay, shown at the bottom left. It learns from a replayed sample that is retrieved from the buffer, denoted as 
$S'': \left(\mathcal{X}''_{N,T}, \mathcal{X}''_{N,T'}, \hat{b}'', \hat{p}'', P_{S''}\right)$, shown at the bottom right.  
(4) learnable variable and parameter vectors $\hat{b}$, $\hat{p}$, along with their corresponding binary masks $b$, $p$, shown at the top right.  
(5) adjacency matrix $\hat{A}$, shown at the top right. In (1), (2) and (3) of the figure, we illustrate the temporal embeddings $E_N^{\mathrm{temp}}$, the feature embedding $E_N^{f}$, and the node embeddings $E_N^{node}$ (gray blocks); spatial or temporal attentions (orange block in (1)); parameter-wise masked attentions (orange blocks in (2) and (3)); and dynamic extrapolation (green blocks).
}

\label{fig:USS}
\end{figure*}

\subsubsection*{\textbf{From STMF to STCV}} 
The proposed STMF model is designed to forecast future values of all time series in $N$ based on their past values. It relies on dense neural networks, i.e., embeddings and attention mechanisms, and requires all variables in the model input.

We regard STCV as a variable reduction problem and produce its model from the pretrained STMF model by iteratively pruning insignificant variables to select a final desired set $M$ of variables. With a reduced number of variables, we also prune insignificant dimensions of model parameters to reduce the model size and thus improve its inference efficiency. Furthermore, reducing input variables and parameter dimensions not only lowers computational cost but also effectively shrinks the combinatorial search space for optimal model configurations, thus enabling more efficient learning~\cite{frankle2018lottery}.

Below we propose a new learning framework for STCV, called \textit{Variable-Parameter Iterative Pruning (VIP)}, where each iteration jointly prunes less-important variables and parameters and the process ends after a budgeted number $m$ of variables are reached.

\subsubsection*{\textbf{Overview}}
VIP jointly tackles the challenges of variable selection and model compression in the STCV setting. It progressively reduces the number of input variables and model parameters through an iterative masking strategy. VIP is composed of the three tightly integrated components:

(1) {\it Masked Variable-Parameter Pruning.} We use learnable masks over variables and attention parameters to identify and eliminate less-informative inputs and model dimensions by quantile-based thresholding. Such a pruning mechanism is illustrated on the left side of Fig.~\ref{fig:USS}.

(2) {\it Dynamic Extrapolation.} We define a unified training objective that integrates forecast loss, replay loss, and stochastic regularization over mask vectors to balance convergence and exploration. We enable dynamic extrapolation by propagating the features learned from chosen variables in $M$ to missing variables in $M'$  via the node embeddings, as shown in the green blocks of Fig.~\ref{fig:USS}.

(3) {\it Prioritized Variable-Parameter Replay.} To alleviate catastrophic forgetting induced by dynamic pruning, we employ a replay buffer that retains previously encountered low-loss samples, prioritizing those deemed more informative. This mechanism is depicted in the bottom-right corner of Fig.~\ref{fig:USS}.

Together, these components form a cohesive learning framework that jointly optimizes variable selection and parameter sparsity, enabling scalable and accurate spatio-temporal forecast.

\subsubsection*{\textbf{Masked Variable-Parameter Pruning}}
We first introduce variable mask and parameter mask. We then explain how to perform variable-parameter pruning.

{\it Variable Mask}. We use a binary variable mask $b \in \{0,1\}^{n}$ to specify variable selection: variable $i$ is selected iff $b_i = 1$, where $b_i$ is the $i$th masking bit of $b$, $\forall i \in [0, n)$. Given a value matrix $\mathcal{X}_{N,T}$ and a variable mask $b$, we define $\mathcal{X}_{N,T}[b]$ as a sub-matrix with only those time series $\mathcal{X}_{i,T}$ whose corresponding masking bits in $b$ are 1's, i.e., $b_i = 1$, where $[b]$ is a {\it masking operator}: When it is applied to a matrix of size $n \times ...$, it selects the rows whose masking bits are ones; when it is applied to a vector of size $n$, it selects the elements whose masking bits are ones.  For another example, given an adjacency matrix $A$ and a variable mask $b$, $A[b]$ is a sub-adjacency matrix with only those rows (variables) whose mask bits in $b$ are 1's. 

{\it Parameter Mask}. Given the set $\Theta$ of parameters in an STMF model, with each of its attentions having $q$ dimensions of parameters, and a binary parameter mask $p \in \{0,1\}^{q}$ that is applied to all attentions, we define $\Theta[p]$ as a subset of parameters with only those dimensions (in each attention) whose mask bits in $p$ are 1's, i.e., $p_j = 1$, where $p_j$ is the $j$th masking bit in $p$. In the sequel, this parameter masking operation $[p]$ will also be applied to other parameter sets of $q$ dimensions. 

{\it Iterative Pruning}. Given a pre-trained STMF model $F(\mathcal{X}_{N,T}, A, \Theta)$, we will adopt $F(\mathcal{X}_{N,T}[b], A[b], \Theta[p])$ as an STCV model that forecasts the future values of all variables, but with fewer input variables $\mathcal{X}_{N,T}[b]$ and fewer model parameters $\Theta[p]$, where $b$ and $p$ are iteratively optimized.

Starting from the initial masks of $b^0$ and $p^0$ whose bits are all ones, our task is to learn $b^k$ and $p^k$ together iteratively, where $k$ is the number of iterations, such that the final masks, denoted as $b^*$ and $p^*$, achieve the best performance for STCV, where $\|b^*\|_0 = m$, and $\|p^*\|_0 = q'$ for a desired target number $q'$ of parameter dimensions in each attention, with $q' < q$. 

Let $\hat{b} \in \mathbb{R}^n$ be a learnable variable importance vector, and $\hat{p} \in \mathbb{R}^q$ be a learnable parameter importance vector. These vectors are optimized jointly with the model parameters during training. The binary variable mask $b^k$ and parameter mask $p^k$ at the $k$-th iteration are derived from $\hat{b}$ and $\hat{p}$ via quantile-based thresholding.

The vector $\hat{b}$ is initialized based on the normalized graph structure: $\hat{b} = \left[\sum_{j=1}^n \hat{A}_{0j}, \ldots, \sum_{j=1}^n \hat{A}_{nj}\right]^\top$, where $\hat{A} = \tilde{D}^{-\frac{1}{2}} (A + I) \tilde{D}^{-\frac{1}{2}}$ is the symmetrically normalized adjacency matrix with self-loops. The $i$-th value of $\hat{b}$ is denoted $\hat{b}_i$, for $i \in [0, n)$. The vector $\hat{p}$ is initialized from a standard normal distribution, with $\hat{p}_j$ denoting the $j$-th element, for $j \in [0, q)$.

The binary masks $b^k$ and $p^k$ are updated as follows:
\begin{align}
b^k_i &= 
\begin{cases}
0, & \text{if } |\hat{b}_i| \leq \operatorname{Quantile}(|\hat{b}[b^{k-1}]|, r_b) \\
1, & \text{otherwise}
\end{cases} \label{eq:indi data} \\
p^k_j &= 
\begin{cases}
0, & \text{if } |\hat{p}_j| \leq \operatorname{Quantile}(|\hat{p}[p^{k-1}]|, r_p) \\
1, & \text{otherwise}
\end{cases} \label{eq:indi param}
\end{align}

Here, $[\,\cdot\,]$ denotes the masking operator that selects elements whose masking bits are 1s. The operator $|\,\cdot\,|$ takes element-wise absolute values, and $\operatorname{Quantile}(x, \beta)$ denotes the $\beta$-quantile of vector $x$. The scalars $r_b$ and $r_p$ are the pruning rates for variable and parameter masks, respectively.

From Eq. \eqref{eq:indi data}-\eqref{eq:indi param}, after the $k$-th iteration, we will retain $\| b^k \|_0$ variables, where $\| b^k \|_0 =max\{\lfloor n \times (1-r_b)^k \rfloor, 1\}$, and retain $\| p^k \|_0$ dimensions of parameters in each attention, where $\| p^k \|_0 = max\{\lfloor q \times (1-r_p)^k \rfloor, 1\}$. 

Consider a value matrix of all variables, $\mathcal{X}_{N,T} \in \mathbb{R}^{n \times l}$. With the retained variable mask $b^k$, the masked value matrix is $\mathcal{X}_{N,T}[b^k] \in \mathbb{R}^{\| b^k \|_0 \times l}$. From Eq. \eqref{eq:feature emb}-\eqref{eq:cat emb} in the STMF model, we can calculate its aggregated feature $E_N^{all}[b^k] \in \mathbb{R}^{\| b^k \|_0 \times l \times q}$.

To effectively retain the model's learning power while reducing the model parameters, we design a parameter-wise masked attention, which is shown in the orange blocks of Fig. \ref{fig:USS}. It applies the learnable parameter vector $\hat{p}$ and the binary parameter mask $p^k$ to each attention in Eq. \eqref{eq:qkv} and \eqref{eq:score} as follow:
\begin{align}
    \hat{Q}^t&=\left(E_N^{all}[b^k]\right)\left(W_Q^t[p^k]\right)\label{eq:sparse qkv}\\
    \hat{K}^t&=\left(E_N^{all}[b^k]\right)\left(W_K^t[p^k]\right)\\
    \hat{V}^t&=\hat{p}\odot \left(E_N^{all}[b^k]W_V^t \right) \\ 
    H_{N,T}^t[b^k]&=Softmax\left(\frac{\hat{Q}^t(\hat{K}^t)^\top}{\sqrt{d_t}}\right)\hat{V}^t \label{eq:sparse score}
\end{align}
where $W_Q^t[p^k], W_K^t[p^k] \in \mathbb{R}^{q \times \| p^k \|_0}$ and $\odot$ is the element-wise product. Spatial attentions are handled similarly except that transpose is performed on the input, i.e., $\hat{K}^t$.

\subsubsection*{\textbf{Dynamic Extrapolation}}

To ensure accurate forecast, we must propagate learned representations from the variables that are selected to the variables that are not selected. We design a dynamic extrapolation mechanism, which is shown in the green blocks of Fig. \ref{fig:USS}, that constructs a global information bridge between variables in $M$ and those in $M'$. It operates in two stages: (1) computing a soft similarity-based attention matrix over variable embeddings, and (2) fusing it with the original masked adjacency matrix to allow knowledge transfer across all variables.

\paragraph{Step 1: Similarity-based Extrapolation Attention.}
We first compute a global attention matrix $\hat{B} \in \mathbb{R}^{m \times n}$ based on node embedding similarity:
\begin{align}
\hat{B} &= \sigma\left(FC(E_N^{node}[b^k]) \cdot FC(E_N^{node})^\top\right)
\end{align}
where $FC(\cdot)$ is a shared learnable linear projection. The activation $\sigma(\cdot)$ is the GeLU function~\cite{hendrycks2016gaussian}. This attention matrix $\hat{B}$ serves as a dense bridge from the variables selected by mask $b^k$ to all variables.

\paragraph{Step 2: Fusion with Masked Adjacency.}
We then combine $\hat{B}$ with the masked adjacency matrix to form a final extrapolated adjacency:

\begin{align}
\hat{A}' &= \hat{b}[b^k] \odot \hat{A}[b^k] + \hat{B}
\end{align}

Here, $\hat{A}[b^k]$ is the symmetrically normalized adjacency matrix for the selected variables, and $\hat{b}[b^k]$ acts as a learned variable-level weight. This fused matrix $\hat{A}' \in \mathbb{R}^{m \times n}$ enables spatial message passing from the selected variables to all variables.

\paragraph{Step 3: Feature Propagation.}
The extrapolated representation for all variables is computed as:

\begin{align}
H_{N,T} = \hat{A}'^\top \cdot H_{N,T}[b^k]
\label{eq:transfer}
\end{align}

This mechanism allows latent features from the selected variables to influence other  variables via learned attention, enhancing forecast quality under sparse input.

Finally, an MLP is used to map the high-dimensional representation $H_{N,T}$ to the forecast output $\hat{\mathcal{X}}_{N,T'}$ as in Eq. \eqref{eq:final fc}, and the loss function is given below
\begin{align}
\mathcal{L}_{main}(\cdot,\cdot) = \mathcal{L}\left(F(\mathcal{X}_{N,T}[b^k],A[b^k],\Theta[p^k]),\mathcal{X}_{N,T'}\right).\label{eq:main loss}
\end{align}

\subsubsection*{\textbf{Prioritized Variable-Parameter Replay}}
Within the $k$th iteration of VIP, mask $b^k$ for variable selection and mask $p^k$ for parameter selection will be updated together with the learnable vectors $\hat{b}$ and $\hat{p}$ during training.  If the two masks differ significantly after each update, the model might not retain the significant patterns from the previous training, which is a phenomenon known as catastrophic forgetting \cite{kirkpatrick2017overcoming}. The experience replay (ER) methods \cite{zhang2017deeper}, which store some historical samples in a buffer and replay them to the current model using a first-in-first-out strategy, can mitigate the problem of catastrophic forgetting. Furthermore, the prioritized experience replay method \cite{schaul2015prioritized} assigns priority to the samples for replay and achieves better performance.

Motivated from the prior ER work, we introduce the prioritized variable-parameter replay method (PVR) to mitigate catastrophic forgetting in our proposed VIP. After a new sample is applied in training, PVR stores this sample $S: \left(\mathcal{X}_{N,T},\mathcal{X}_{N,T'}, \hat{b},\hat{p}, P_{S}\right)$ in a buffer $\mathcal{R}$, which can keep up to  $\|\mathcal{R}\|$ samples, where the priority $P_{S}$ of the current sample $S$ is given as
\begin{align}
P_{S} &=\frac{1}{\mathcal{L}_{main}(\cdot,\cdot)}\label{eq:store sample}.
\end{align}
Intuitively, in the context of VIP where variable-parameter pruning can cause learned patterns to be pruned from the model, low-loss samples should be given priorities to combat catastrophic forgetting. This assumption is supported by experimental evaluation in Section~\ref{subsec:replay}, confirming that samples yielding lower $\mathcal{L}_{main}$ help retain significant patterns. 

When inserting a new sample to $\mathcal{R}$, if $\mathcal{R}$ is full, PVR will probabilistically select an existing sample from $\mathcal{R}$, replay it, and then remove it. The probability for a sample $S' \in  \mathcal{R}$ to be selected is 
\begin{align}
Prob(S') &= \frac{(P_{S'})^\alpha}{\sum_{\forall S' \in \mathcal{R}} (P_{S'})^\alpha}\label{eq:sample rate}
\end{align}
where $\alpha$ is the hyperparameter. 
Let $S'':\left(\mathcal{X}''_{N,T}, \mathcal{X}''_{N,T'},{\hat{b}''},{\hat{p}''}, P_{S''}\right)$ denote the sample selected for replay. Its loss function is 
\begin{align}
\mathcal{L}_{replay}(\cdot,\cdot) = \mathcal{L} \left( F(\mathcal{X}''_{N,T}[b^{''k}], A[b^{''k}],\Theta[p^{''k}]),\mathcal{X}''_{N,T'} \right) \label{eq:replay loss}
\end{align}
where the values of $b^{''k}$ and $p^{''k}$ are obtained from ${\hat{b}''}$ and ${\hat{p}''}$ through Eq. \eqref{eq:indi data}-\eqref{eq:indi param} by replacing $\hat{b}''$ for ${\hat{b}}$ and $\hat{p}''$ for ${\hat{p}}$.

\begin{algorithm}[tb]
   \caption{$k$-th iteration of variable-parameter iterative pruning in model learning}
   \label{alg:USS}
\begin{algorithmic}[1]
   \STATE \textbf{Input:} Value matrix $\mathcal{X}_{N,T}$, adjacency matrix $A$, STMF model $F(\cdot,A,\Theta)$, trainable variable mask $\hat{b}$ and parameter mask $\hat{p}$, buffer $\mathcal{R}$, learning rate $\lambda$
   \STATE \textbf{Output:} Sparsified variable mask $b^k$, parameter mask $p^k$, model parameters $\Theta$
   \STATE $\mathcal{R} \gets \{\}$
   \REPEAT

   \STATE \quad\textit{// Compute binary masks for current sample}
   \STATE Compute variable mask $b^k$ and parameter mask $p^k$ from $\hat{b}$ and $\hat{p}$ by Eq.~\eqref{eq:indi data}-\eqref{eq:indi param}
   
   \STATE \quad\textit{// Forward pass on current sample}
   \STATE Forward $F(\mathcal{X}_{N,T}[b^k], A[b^k], \Theta[p^k])$ to compute the main loss $\mathcal{L}_{\text{main}}(\cdot,\cdot)$ by Eq.~\eqref{eq:main loss}
   
   \STATE \quad\textit{// Replay sample selection and loss computation}
   \STATE Select a replay sample $(\mathcal{X}''_{N,T}, \mathcal{X}''_{N,T'}, \hat{b}'', \hat{p}'')$ from $\mathcal{R}$ using Eq.~\eqref{eq:store sample} and Eq.~\eqref{eq:sample rate}
   \STATE Compute variable mask $b^{''k}$ and parameter mask $p^{''k}$ from $\hat{b}''$ and $\hat{p}''$ by Eq.~\eqref{eq:indi data}-\eqref{eq:indi param}
   \STATE Forward $F(\mathcal{X}''_{N,T}[b^{''k}], A[b^{''k}], \Theta[p^{''k}])$ to compute the replay loss $\mathcal{L}_{\text{replay}}(\cdot,\cdot)$ by Eq.~\eqref{eq:replay loss}
   
   \STATE \quad\textit{// Total loss and gradient update}
   \STATE Obtain total loss $\mathcal{L}_{\text{sum}}(\cdot,\cdot)$ by Eq.~\eqref{eq:sum loss}
   \STATE Backpropagate to update $\Theta \gets \Theta - \lambda \nabla_{\Theta} \mathcal{L}_{\text{sum}}(\cdot,\cdot)$
   \STATE Update $\hat{b} \gets \hat{b} - \lambda \nabla_{\hat{b}} \mathcal{L}_{\text{sum}}(\cdot,\cdot)$
   \STATE Update $\hat{p} \gets \hat{p} - \lambda \nabla_{\hat{p}} \mathcal{L}_{\text{sum}}(\cdot,\cdot)$
   
   \STATE \quad\textit{// Store current sample into replay buffer}
   \STATE Insert current sample $(\mathcal{X}_{N,T}, \mathcal{X}_{N,T'}, \hat{b}, \hat{p}, P_S)$ into $\mathcal{R}$

   \UNTIL{all batches in epoch are processed}
\end{algorithmic}
\end{algorithm}

\subsubsection*{\textbf{Co-optimized Variable-Parameter Loss}}
In practice, there is a balance between adjusting and stabilizing the selected variables and model parameters. On the one hand, adjusting more aggressively introduces more combinations of selection. On the other hand, it makes the selection harder to converge. Stabilizing the masks aggressively accelerates convergence, but it may lead to suboptimal forecast performance.

To address this issue, we introduce a new {\it co-optimized variable-parameter loss} function $\mathcal{L}_{sum}$. This loss consists of three components: the main loss $\mathcal{L}_{main}$, which aims to find the optimal selection of of variables and model parameters; the replay loss $\mathcal{L}_{replay}$, which helps the model retain knowledge of previous samples; the random regularization of $\hat{b}$ and $\hat{p}$, which encourages sparse and diverse pruning decisions during training, acting as a stochastic exploration mechanism to avoid premature convergence to suboptimal masks. The co-optimized variable-parameter loss is defined as
\begin{align}
\mathcal{L}_{sum}(\cdot,\cdot) = &\mathcal{L}_{main}(\cdot,\cdot)+\gamma_1\mathcal{L}_{replay}(\cdot,\cdot) +\nonumber\\&\gamma_2\|\hat{b}[r_1]\|_1+\gamma_3\|\hat{p}[r_2]\|_1\label{eq:sum loss}
\end{align}
where the trainable variable vector $\hat{b}$ is optimized on a subset of variables chosen by a random mask $r_1 \in \{0,1\}^{n}$ with $\|r_1\|_0$ being a small number, the trainable parameter vector $\hat{p}$ is optimized on a subset of parameter dimensions chosed by another random mask $r_2 \in \{0,1\}^{q}$ with $\|r_2\|_0$ being a small number, and $\gamma_1$,   $\gamma_2$, and $\gamma_3$ are the hyperparameters.

% \subsubsection*{\textbf{Complexity Analysis}}
% The inference time complexity of our VIP model is $\mathcal{O}\left(  m\times l \times q + L \times m\times l \times q' \times q + m \times n \times l \right)$, where $m$ is the desired number of variables to selected, $q'$ is the desired number of parameter dimenstions in each attension to retain, $\mathcal{O}\left(m\times l \times q  \right)$ is the time complexity of embeddings, $L$ is the total number of attention layers, $\mathcal{O}\left(m\times l \times q' \times q\right)$ is the time complexity of each parameter-wise masked attention, and $\mathcal{O}\left(m \times n \times l\right)$ is the time complexity of dynamic extrapolation. The memory complexity is $\mathcal{O}\left(m \times l \times q +  L \times q' \times q + n + L \times q \right)$, where $\mathcal{O}\left(m \times l \times q \right)$ is the memory complexity of embeddings, $\mathcal{O}\left(q' \times q \right)$ is the memory complexity of the dynamic extrapolation,  $\mathcal{O}\left(n\right)$ is the memory complexity of the learnable variable vector $\hat{b}$, and $\mathcal{O}\left(q\right)$ is the memory complexity of the learnable parameter vector $\hat{p}$.

\subsubsection*{\textbf{Complexity Analysis}}

We analyze the inference complexity of our VIP model and compare it to the base STMF model. Let $m$ be the number of chosen variables, $n$ the total number of variables, $l$ the length of the time period in the model input (output), $L$ the number of attension layers, $q$ the original number of attention dimensions, and $q'$ the after-prune number of attention dimensions. 

\paragraph{Time Complexity.} The total inference time complexity of the VIP model is:
\begin{align*}
\mathcal{O}\left(
    m \cdot l \cdot q
    + L \cdot m \cdot l \cdot q' \cdot q
    + m \cdot n \cdot l
\right)
\end{align*}
The three terms correspond to: (1) variable-wise feature embedding, (2) attention computation, and (3) extrapolation from the chosen variables to all variables. In contrast, the original STMF model has a time complexity of:
\begin{align*}
\mathcal{O}\left(
    n \cdot l \cdot q
    + L \cdot n \cdot l \cdot q^2
\right)
\end{align*}
By reducing the number of input variables from $n$ to $m$, and attention dimensions from $q$ to $q'$, our model achieves significant computational savings, especially when $m \ll n$ and $q' \ll q$.

\paragraph{Memory Complexity.} The total memory complexity of the STCV model is:
\begin{align*}
\mathcal{O}\left(
    m \cdot l \cdot q
    + L \cdot q' \cdot q
    + n
    + L \cdot q
\right)
\end{align*}
where the terms corresponds to the embeddings, the pruned attention layers, and the trainable masks, $\hat{b}$ and $\hat{p}$. In comparison, the memory complexity of the STMF model is:
\begin{align*}
\mathcal{O}\left(
    n \cdot l \cdot q
    + L \cdot q^2
\right)
\end{align*}
The STCV model saves both space and time by working with a sparse set of input variables and a sparse set of parameters during inference, while still forecasting for all $n$ variables.

\section{Trace-based Experimental Evaluation}

\newcolumntype{B}{!{\vrule width 1pt}}
\renewcommand{\arraystretch}{1.4}

\begin{table*}[t]
\centering
\caption{Performance comparison over five datasets, where $n$ is the total number of locations and $m$ is the number of chosen locations.  }
\label{tab:all_performance_comparison}
\resizebox{\textwidth}{!}{
\begin{tabular}{|c|Bc|c|c|Bc|c|c|Bc|c|c|Bc|c|c|Bc|c|c|B}
\Hline
\multirow{3}{*}{\textbf{Models}} & \multicolumn{3}{c|}{\textbf{PEMS04}} & \multicolumn{3}{c|}{\textbf{PEMS08}} & \multicolumn{3}{c|}{\textbf{PEMSBAY}} & \multicolumn{3}{c|}{\textbf{METRLA}} & \multicolumn{3}{c|}{\textbf{AQI}} \\ \cline{2-16} 
& \multicolumn{15}{c|}{\textbf{deployment ratio $m/n=10\%$, sparsity ratio $1- m/n=90\%$}}\\ \cline{2-16} 
                       & MAE   & RMSE  & MAPE  & MAE   & RMSE  & MAPE   & MAE    & RMSE   & MAPE   & MAE   & RMSE  & MAPE  & MAE   & RMSE  & MAPE  \\ \Hline
Matrix Factorization  & 92.13 & 141.41 & 124.29 & 75.80 & 131.17 & 82.16 & 6.51 & 14.29 & 21.09 & 15.61 & 33.39 & 48.51 & 67.63 & 115.43 & 130.18 \\\hline
STID                   & 94.72 & 123.30 & 149.68 & 81.64 & 113.20 & 86.54 & 3.78   & 8.65   & 10.07  & 6.24  & 12.06 & 23.17 & 25.97 & 40.28 & 62.41 \\\hline
DSformer               & 40.31 & 59.18  & 25.62  & 38.79 & 55.34  & 32.61  & 3.72   & 9.06   & 10.23  & 5.02  & 10.15 & 17.69 & 23.17 & 35.72 & 57.31 \\\hline
STGODE  & 30.91 & 48.25 & 25.59 & 26.73 & 41.20 & 23.20 & 3.29 & 8.22 & 9.76 & 4.48 & 8.49 & 14.79 & 23.78 & 38.69 & 47.10 \\\hline
STSGCN                & 34.42 & 50.51  & 22.09  & 28.51 & 46.31  & 17.94  & 3.23   & 7.73   & 8.41   & 4.39  & 8.45  & 12.61 & 25.61 & 40.74 & 59.96 \\\hline
MegaCRN                & 36.14 & 52.17  & 23.42  & 34.52 & 52.75  & 24.64  & 3.54   & 8.25   & 9.23   & 4.58  & 8.72  & 13.54 & 22.61 & 35.74 & 53.28 \\\hline
ST-SSL  & 28.89 & 44.11 & 19.07 & 24.87 & 40.87 & 17.46 & 3.05 & 6.91 & 7.86 & 3.96 & 7.98 & 11.82 & 18.98 & 32.97 & 38.59 \\\hline
TESTAM  & 29.11 & 44.13 & 19.11 & 24.81 & 38.43 & 15.43 & 2.85 & 6.91 & 7.80 & 3.85 & 7.79 & 11.25 & 16.98 & 32.12 & 37.62 \\\hline
STAEFormer  & 28.11 & 43.32 & 18.35 & 24.31 & 39.25 & 15.63 & 2.72 & 6.58 & 6.70 & 4.07 & 8.43 & 12.05 & 17.47 & 32.60 & 39.09 \\\hline
STAEFormer + IGNNK  & 27.89 & 42.97 & 18.21  & 24.99 & 39.15 & 15.32 & 2.80 & 6.90 & 6.85 & 4.32 & 8.78 & 12.21 & 17.32 & 33.32 & 40.12 \\\hline
STAEFormer + STGNP  & 28.10 & 42.40 & 18.39 & 24.37 & 39.39 & 15.91 & 2.73 & 6.60 & 6.79 & 4.19 & 8.58 & 12.29 & 17.59 & 34.15 & 41.61 \\\hline
DCRNN+GPT4TS           & 31.42 & 46.17  & 21.27  & 31.78 & 46.22  & 22.96  & 3.31   & 7.69   & 8.82   & 4.29  & 8.31  & 12.18 & 21.78 & 34.28 & 50.64 \\\hline
DFDGCN+TimesNet        & 30.98 & 45.93  & 21.43  & 30.46 & 45.83  & 21.59  & 3.27   & 7.15   & 8.34   & 4.34  & 8.33  & 12.24 & 21.06 & 33.19 & 51.28 \\\hline
MTGNN+GRIN             & 30.61 & 45.88  & 21.04  & 29.15 & 42.79  & 21.33  & 3.22   & 7.26   & 8.45   & 4.25  & 8.29  & 12.14 & 20.02 & 34.15 & 50.23 \\\hline
FourierGNN+GATGPT      & 32.16 & 48.83  & 23.65  & 29.11 & 41.44  & 20.61  & 3.28   & 7.44   & 9.07   & 4.33  & 8.37  & 12.28 & 20.18 & 34.17 & 50.94 \\\hline
LGnet                  & 36.29 & 55.12  & 23.74  & 35.48 & 52.06  & 25.71  & 3.67   & 8.92   & 9.83   & 4.79  & 9.15  & 14.38 & 23.24 & 35.54 & 57.18 \\\hline
GC-VRNN                & 30.06 & 45.12  & 20.43  & 28.46 & 41.98  & 20.54  & 3.12   & 7.32   & 7.94   & 4.32  & 8.35  & 12.29 & 19.24 & 32.91 & 48.73 \\\hline
TriD-MAE               & 29.54 & 44.23  & 20.05  & 26.18 & 39.25  & 16.43  & 3.02   & 7.02   & 7.63   & 4.11  & 8.22  & 11.92 & 18.04 & 32.68 & 45.76 \\\hline
BiTGraph               & 28.71 & 43.37  & 19.08  & 25.01 & 39.06  & 16.18  & 2.83   & 6.75   & 7.42   & 3.91  & 8.03  & 11.78 & 17.06 & 31.85 & 38.17 \\\hline
GinAR               & 28.20 & 42.82  & 18.31  & 24.83 & 38.87  & 15.82  & 2.77   & 6.43   & 6.94   & 3.87  & 7.84  & 11.25 & 16.83 & 30.97 & 36.30 \\\hline
MaxConnectivity+GinAR &  28.11 &  42.65 &  18.14 &  24.89 &  39.07 &  15.99 & 2.84 &  6.80 & 7.23 &  3.93 &  8.09 &  12.21 &  16.97 &  31.35 & 37.89 \\\hline
MaxValue+GinAR &  28.04 &  42.43  &  18.67 &  24.59 &  38.27 &  15.95 & 2.67 &  6.30 & 6.73 &  3.78 &  7.83 &  11.20 &  16.71 &  31.54 & 37.75 \\\hline
Grid+GinAR &  28.87 &  42.31 &  18.06 &  24.70 &  38.84 &  15.74 & 2.78 &  6.59 & 6.93 &  3.90 &  7.99 &  11.36 &  16.89 &  31.14 & 37.52 \\\hline
Metaheuristics+GinAR &  30.54 &  46.83  &  21.12 &  26.98 &  41.62 &  17.81 & 3.10 &  7.19 & 8.22 &  4.51 &  8.93 &  13.45 &  19.37 &  34.97 & 39.33 \\\hline
NIP+Pre &  22.99 &  38.51 &  16.23 &  18.62 &  31.98 &  13.01 & 2.43 &  4.89 & 5.79 &  3.80 &  7.91 &  11.05 &  16.23 &  30.03 & 35.92 \\\hline
\textbf{VIP wo. pretrain} &  23.99 &  39.27 &  16.05 &  19.87 &  32.73 &  13.19 &  2.44 &  4.99 &  5.80 &  3.52 &  7.14 &  10.20 &  16.98 &  31.34 &  35.21 \\\hline
\textbf{VIP w. pretrain} &  22.93 &  38.26 &  16.06 &  18.59 &  31.51 &  12.63 & 2.38 &  4.76 & 5.64 &  3.64 &  7.68 &  10.94 &  15.89 &  29.22 & 35.01 \\\hline

\end{tabular}
}
\end{table*}

\subsection{Settings and Datasets} \label{sec:setting}

\begin{table}[t]
\begin{center}
\captionsetup{font=small}
\caption{ Basic statistics of the datasets used in our experiments.}
 \aboverulesep=0ex
 \belowrulesep=0ex
\setlength{\tabcolsep}{3pt}
\renewcommand{\arraystretch}{1.2}
\large
\centering
\rmfamily
\label{tab:Data_Statistics}
\resizebox{\linewidth}{!}{
\begin{tabular}{ c | c  | c | c  | c | c} 
\toprule
\textbf{Dataset} & \textbf{AQI} & \textbf{PEMS04}  & \textbf{PEMS08} & \textbf{PEMS-BAY} & \textbf{METR-LA} \\
\midrule
Area & China & \multicolumn{4}{c}{CA, USA}\\ 
\midrule
\multirow{2}{*}{Time Span} & 1/1/2015 -  & 1/1/2018 -   & 7/1/2016 - & 1/1/2017 - & 3/1/2012 -\\ 
& 12/31/2022 & 2/28/2018  & 8/31/2016 & 5/31/2017 & 6/30/2012 \\
\midrule
Time Interval & 1 hour & \multicolumn{4}{c}{5 min}\\ 
\midrule
Number of Locations, $n$ & 350 & 307   & 170 & 325 & 207\\ 
\midrule
Number of Time Intervals & 59,710 & 16,992   & 17,856 & 52,116 & 34,272\\ 
\bottomrule
\end{tabular}}
\end{center}
\end{table} 
\textbf{Datasets: } We use five widely used public datasets, including four road traffic datasets (METR-LA, PEMSBAY, PEMS04 and PEMS08) and an air quality dataset (AQI) \footnote{The datasets are provided in the GinAR github repository at \url{https://github.com/GestaltCogTeam/GinAR/}}, for the road traffic forecast application and the air quality forecast application in the introduction, respectively. The details of data statistics are shown in Table \ref{tab:Data_Statistics}. We split METR-LA, PEMSBAY dataset into three subsets in 7:1.5:1.5 ratio for training, validation, and test and 3:1:1 for the other datasets. 

These datasets are mapped to the STCV problem as follows: Each of the $n$ locations in a dataset corresponds to a variable whose values are the time series of traffic (air quality) measurements at the location. After collecting the training/validation data subsets with cheap, temporary sensors from $n$ locations (i.e., variables), we select $m$ locations for permanent sensor deployment in the test data subsets for evaluation. These $m$ {\it chosen locations} correspond to the chosen variables in the model input; the other $(m' = n-m)$ {\it unchosen locations} correspond to the missing variables not in the model input. 

We define the sensor {\it deployment ratio} as $\frac{m}{n}$ and the deployment {\it sparcity ratio} as $1 - \frac{m}{n}$. In the application context of our evaluation description, we will use the term {\it location} for {\it variable} in the general STCV definition. 

\textbf{Hyperparameters: } $m = 0.1 n$ by default; hence, the deployment ratio is 10\% and 90\% by default --- note that different settings of $m$ and deployment (sparcity) ratios will be studied. $l = l' = 12$ for one hour forecast in the datasets. We use the Mean Absolute Error (MAE) as the loss function. All experiments are conducted on a server equipped with an AMD EPYC 7742 64-Core Processor @ 2.25 GHz, 500 GB RAM, and an NVIDIA A100 GPU with 80 GB memory.

We perform a systematic grid search over other hyperparameters, as summarized in Table~\ref{tab:hyperparam_summary}. These hyperparameters are categorized into two groups: primary and secondary. Each hyperparameter is tested across four candidate values, and we explore all combinations within each group independently.

The primary group includes the node pruning rate $r_b$, parameter pruning rate $r_p$, size $|\mathcal{R}|$ of the replay buffer, priority coefficient $\alpha$, learning rate, and batch size.
The secondary group contains the loss-related coefficients $\gamma_1, \gamma_2, \gamma_3$, the number $|r_1|_0$ of selected elements in the node vector, and the number $|r_2|_0$ of selected elements in the parameter vector, as defined in Eq.~\ref{eq:sum loss}.

For the remaining STMF hyperparameters, we follow the default settings established in prior work~\cite{liu2023spatio}, namely: $q=152$, $d=24$, $d_{\text{tod}}=24$, $d_{\text{dow}}=24$, $d_v=80$, and $L=6$.

\begin{table}[ht]
\centering
\caption{hyperparameter settings with the deployment rate of $10\%$ on all datasets, where the bold values represent the best setting}
\label{tab:hyperparam_summary}
\resizebox{0.80\columnwidth}{!}{\begin{tabular}{|c|c|c|c|c|c|}
\hline
\textbf{Group} & \textbf{Hyperparameter} & \textbf{Value 1} & \textbf{Value 2} & \textbf{Value 3} & \textbf{Value 4}\\
\hline
\multirow{6}{*}{primary } & $r_b$ & 0.05 & \textbf{0.10} & 0.20 & 0.40\\ \cline{2-6}
& $r_p$ & 0.01 & 0.02 & \textbf{0.05} & 0.10\\ \cline{2-6}
& $\|\mathcal{R}\|$ & 288 & 288*3 & \textbf{288*7} & 288*30\\ \cline{2-6}
& $\alpha$ & 0.2 & 0.4 & \textbf{0.6} & 0.8\\ \cline{2-6}
& batch size & 8 & 16 &  \textbf{64} & 128\\ \cline{2-6}
& learning rate & $10^{-2}$ & $\boldsymbol{10^{-3}}$ & $5\times10^{-4}$ & $10^{-4}$\\ \hline
\multirow{5}{*}{secondary} &$\gamma_1$ & 0.1 & 0.5 & \textbf{1.0} & 2.0\\ \cline{2-6}
& $\gamma_2$ & 0.1 & 0.5 & \textbf{1.0} & 2.0\\ \cline{2-6}
& $\gamma_3$ & 0.1 & 0.5 & \textbf{1.0} & 2.0\\ \cline{2-6}
& $\|r_1\|_0$ & 0 & 1 & \textbf{2} & 5\\ \cline{2-6}
& $\|r_2\|_0$ & 0 & \textbf{1} & 2 & 5 \\ \hline
\end{tabular}}
\end{table}

% \begin{table}[ht]
% \centering
% \caption{MAE performance under different hyperparameter settings when the deployment rate at $10\%$ on PEMS08 dataset}
% \label{tab:hyperparam_summary}
% \resizebox{0.95\columnwidth}{!}{\begin{tabular}{|l|c|c|c|c|}
% \hline
% \textbf{Hyperparameter} & \textbf{Value 1} & \textbf{Value 2} & \textbf{Value 3} & \textbf{Value 4}\\
% \hline
% $r_b$ & 18.63 (0.05) & \textbf{18.59 (0.10)} & 18.60 (0.20) & 18.65 (0.20)\\ \hline
% $r_p$ & 18.59 (0.01) & \textbf{18.59 (0.05)} & 18.63 (0.10) & 18.69 (0.20)\\ \hline
% $\|\mathcal{R}\|$ & 18.79 (288) & 18.68 (288*3) & \textbf{18.59 (288*7)} & 18.60 (288*30)\\ \hline
% $\alpha$ & 18.87 (0.2) & 18.69 (0.4) & \textbf{18.59 (0.6)} & 18.72 (0.8)\\ \hline
% Batch size & 18.59 (8) & 18.70 (16) &  \textbf{18.59 (64)} & 18.73 (128)\\ \hline
% Learning rate & 18.62 ($10^{-2}$) & \textbf{18.59 ($10^{-3}$)} & 18.64 ($5\times10^{-4}$) & 18.64 ($10^{-4}$)\\ 
% \hline
% \end{tabular}}
% \end{table}

\textbf{Baselines: } \label{subsec:baselines}The baselines for comparison include spatio-temporal forecast models, such as STGODE \cite{fang2021spatial}, ST-SSL \cite{ji2023spatio}, STID \cite{shao2022spatial}, STSGCN \cite{song2020spatial}, DSformer \cite{yu2023dsformer}, MegaCRN \cite{jiang2023spatio}, STAEFormer \cite{liu2023spatio}, and TESTAM \cite{lee2024testam}; two-stage spatial extrapolation models, which first perform spatio-temporal forecast (using past measurements from the $m$ chosen locations to predict the future measurements at the $m$ locations) and then perform spatial extrapolation (using the predicted future measurements at the $m$ locations to extrapolate the future measurements at all $n$ locations), such as DCRNN\cite{li2018diffusion}+GPT4TS\cite{zhou2023one}, DFDGCN\cite{li2024dynamic}+TimesNet\cite{wu2022timesnet}, MTGNN\cite{wu2020connecting}+GRIN\cite{cini2021filling}, FourierGNN\cite{yi2023fouriergnn}+GATGPT\cite{chen2023gatgpt}, STAEFormer+IGNNK \cite{wu2021inductive} and STAEFormer+STGNP \cite{hu2023graph}; and the spatio-temporal imputation models, such as LGnet \cite{tang2020joint}, GC-VRNN \cite{xu2023uncovering}, TriD-MAE \cite{zhang2023trid}, BiTGraph \cite{chen2023biased} and GinAR \cite{yu2024ginar} on STMF with missing variables, which is most related to our work; and a non-deep-learning Matrix Factorization method \cite{lee2000algorithms}. Adapting the baselines to work for the problem of STCV, following GinAR \cite{yu2024ginar}, we set the values of unchosen variables in their input value matrix to 0s.  

In addition, we extend comparison to cover sensor placement strategies from the wireless sensor network domain and other heuristic approaches by including four two-phase baselines, with the first phase selecting sensor locations \cite{meguerdichian2001coverage,gupta2016genetic,yu2024ginar} and the second phase using GinAR \cite{yu2024ginar} to perform forecast. These baselines are referred to as Grid\cite{meguerdichian2001coverage}+GinAR\cite{yu2024ginar}, Metaheuristics\cite{gupta2016genetic}+GinAR\cite{yu2024ginar}, MaxConnectivity+GinAR\cite{yu2024ginar}, and MaxValue+GinAR\cite{yu2024ginar}, where Grid segments the region (as shown in Fig.~\ref{vis:LA_visual}) into a set of rectangular grids and chooses the one with the highest connectivity within each grid for sensor deployment, Metaheuristics optimize sensor placement using a genetic algorithm (GA)~\cite{holland1992genetic}, MaxConnectivity chooses the locations with the highest connections (node degrees), and MaxValue chooses the locations with the highest values (e.g., traffic rates) on average. Finally, we present a hybrid model, VIP+Pre, where half of the sensor locations are selected by MaxValue (which could be due to a government policy) and then the other half are selected by our proposed VIP. It covers the case where some chosen variables are pre-determined and others can be optimally selected.

For all baselines, we use the recommended hyperparameters from their GitHub repository. Before feeding a value matrix as input to the model, we preprocess the value matrix using z-score normalization, standardizing each value by subtracting the mean flow rate and dividing by the standard deviation.

\textbf{Accuracy Metrics: } We evaluate the forecast accuracy of the proposed work and the baseline models by Mean Absolute Error (MAE), Root Mean Squared Error (RMSE), and Mean Absolute Percentage Error (MAPE).  
Given the ground truth $\mathcal{X}_{i,j}$  and the forecast result $\hat{\mathcal{X}}_{i,j}$, $i \in N$, $j \in T'$, $\text{MAE at time\ } j =  \frac{1}{n}\sum\limits_{i=1}^{n}\left|\mathcal{X}_{i,j} -\hat{\mathcal{X}}_{i,j}\right|$, \ \ \ 
$\text{MAPE at time\ } j = \Big(\frac{1}{n}\sum\limits_{i=1}^{n}\left|\frac{\mathcal{X}_{i,j} -\hat{\mathcal{X}}_{i,j}}{\mathcal{X}_{i,j}}\right| \Big) \times 100\%$, \ \ \ 
and $\text{RMSE at time\ } j =  \sqrt{\frac{1}{n}\sum\limits_{i=1}^{n}\Big(\mathcal{X}_{i,j} -\hat{\mathcal{X}}_{i,j}\Big)^2}$.

\subsection{Forecast Accuracy}\label{sec:forecast acc}
Our VIP model includes two variants: VIP without pretraining, denoted as VIP wo. pretrain, and VIP with pretraining, denoted as VIP w. pretrain. The VIP w. pretrain variant builds upon a pretrained STMF model, as described in Section~\ref{sec:STMF}, where the STMF model is first trained to convergence with the best validation. In contrast, VIP wo. pretrain uses the same base STMF architecture without its parameters being pretrained before VIP training (which will train those parameters). Table~\ref{tab:all_performance_comparison} shows that our VIP w. pretrain consistently outperforms all baseline models in forecast accuracy, for example, achieving $30\%$ improvement by VIP w. pretrain in RMSE against the best baseline result over the PEMSBAY dataset. VIP wo. pretrain performs second best on REMS04, PEMS08 and PEMSBAY, clearly better than all baselines, while the improvement on METRLA and AQI is less pronounced. Yet it takes less time for our VIP wo. pretrain to be trained than most baselines as we will demonstrate in the next subsection. The non-deep-learning matrix factorization method performs poorly probably because it is unable to capture complex spatial-temporal dependencies.  Among all the spatial-temporal models, STAEFormer \cite{liu2023spatio} achieves the best overall performance, benefiting from its embedding techniques (e.g., node embedding, time-of-day and day-of-week embedding). TESTAM \cite{lee2024testam} performs close to STAEFormer, thanks to its informative mixture-of-experts (MOE) \cite{shazeer2017outrageously} architecture. The two-stage spatial-temporal extrapolation models suffer from the error accumulation problem after the first stage of spatial-temporal multivariate forecast.  The spatial-temporal imputation methods substitute zeros for the missing values, which could hurt the forecast accuracy as the substituted zeros and the true zeros (or low values) are mixed in the input. The two-phase baselines, which select sensor locations heuristically and then learn their forecast models based on the selection, underperform when compared to our VIP model, which integrates location selection and forecast training in the same iterative learning process.

\subsection{A Case Demonstration of VIP's Location Selection}\label{sec:case study}
\begin{figure}

        \centering
        \includegraphics[width=0.4\textwidth]{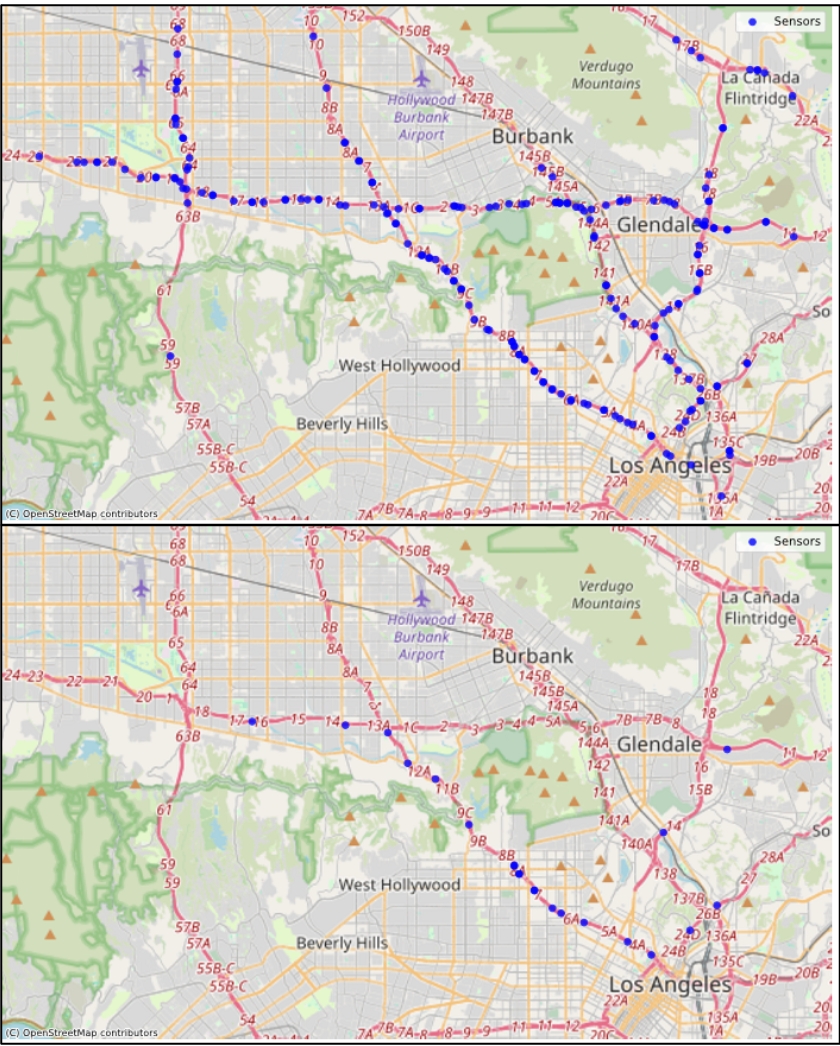}

    \caption{Visualization of sensor locations with METR-LA dataset.}
    \label{vis:LA_visual}
\end{figure}

To help understand the problem of STMF with chosen variables (locations), we use a concrete example to demonstrate qualitatively and visually VIP's location selection in Fig.~\ref{vis:LA_visual}, where we show the full set $N$ of locations of interest in the METRLA dataset on a map in the top plot and the subset $M$ of locations that are selected by VIP in one execution with 10\% deployment ratio in the bottom plot. The figure shows conceptually that while the chosen locations effectively covers (or are close to) major junctions, their density thins out disproportionally along different routes. For example, we observe more locations are chosen along the US-101 corridor (from 24B to 13A to 24) than along the I-5 highway (from 26B to 5 to 152). Such selection bias appears to arise from higher topological connectivity and greater traffic variability along the US-101 route, which connects central Los Angeles to Burbank. These characteristics make the US-101 corridor more informative under our pruning objective. This spatially diverse selection strategy enables more accurate forecasting even under a constrained budget of just 10\% deployment as shown in Table~\ref{tab:all_performance_comparison}.

\subsection{Efficiency Study}

\begin{figure}[htbp]
\centering
\includegraphics[width=0.8\columnwidth]{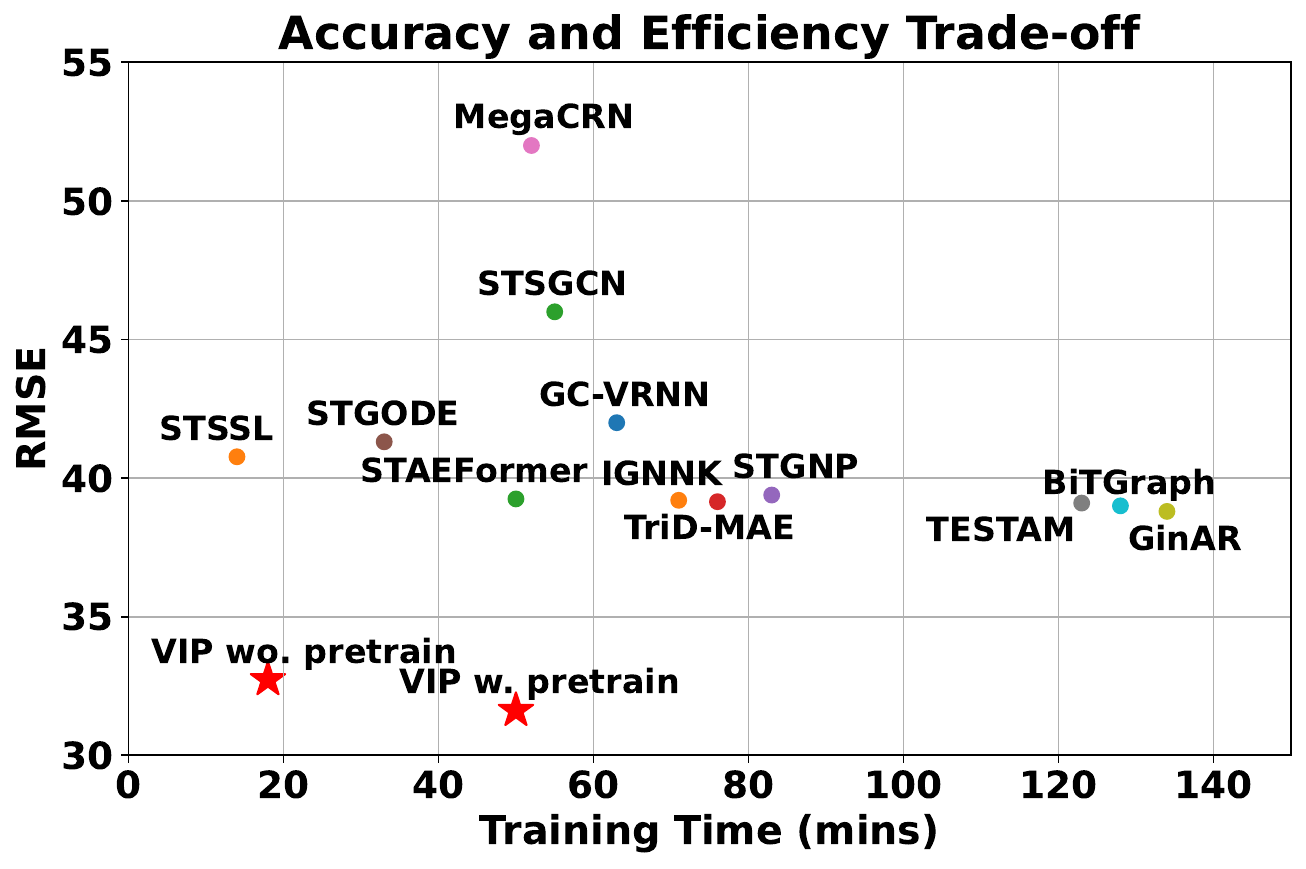}
% \captionsetup{font=small}
\centering
\vspace{-2mm}
\caption{Comparing our VIP models with the baselines in accuracy-efficiency tradeoff, over dataset PEMS08, with a deployment ratio of $m/n=10\%$.
} 
\label{fig:runtime}
\vspace{-2mm}
\end{figure}

Fig. \ref{fig:runtime} compares our VIP models with the baseline models in efficiency-accuracy tradeoff. The batch size is uniformly set to 16. Our models achieve the best RMSE among all models with relatively low computational cost. This dual advantage of high accuracy and efficiency sets our models apart in practical forecast. Note that the training time of VIP w. pretrain includes the training time of its base STMF model. The reason for VIP's efficiency is that we prune locations and parameters with the variable mask and the parameter mask, which reduces both the input size and the network size during training. 

%Most models have a training time complexity of $\mathcal{O}\left( n \times l \times q + L \times n \times l \times q^2 + n^2 \times l \right)$, where $\mathcal{O}\left( n \times l \times q \right)$ represents the complexity of embeddings, $\mathcal{O}\left( L \times n \times l \times q^2 \right)$ represents the complexity of attentions, and $\mathcal{O}\left( n^2 \times l \right)$ represents the complexity of graph operations. These models set the values at missing sensors to 0, thereby retaining $n$ nodes throughout training. In contrast, we treat these nodes as pruned, resulting in $m$ desired nodes (locations), where $m \ll n$. Furthermore, these models maintain a large set of model parameters with dimension at $q$, whereas we retain only $q' \ll q$ in the end.

\subsection{Ablation Study}

\begin{table}[t]
\centering
\caption{Different ablation methods on PEMS08 dataset over the average horizon when the sensor deployment rate $m/n=10\%$.}
\vspace{-3mm}
\resizebox{0.7\columnwidth}{!}{
\begin{tabular}{|c|c|c|c|c|c|c|}
\hline
\multirow{2}{*}{\textbf{Models}} & \multicolumn{3}{c|}{\textbf{wo. pretrain}} & \multicolumn{3}{c|}{\textbf{w. pretrain}}\\ \cline{2-7} 
& MAE & RMSE & MAPE & MAE & RMSE & MAPE\\ 
\hline
VIP & 19.87 & 32.73 & 13.19 & 18.59 & 31.33 & 12.63\\ \hline
VIP wo. extra & 22.10 & 38.42 & 14.39 & 21.19 & 37.78 & 13.47\\ \hline
VIP wo. b & 20.63 & 33.47 & 13.91 & 19.22 & 31.79 & 12.85\\ \hline
VIP wo. p & 20.77 & 33.22 & 15.90 & 18.93 & 31.48 & 13.88 \\ \hline
VIP wo. replay & 20.61 & 33.50 & 15.61 & 19.20 & 32.02 & 13.55\\ \hline
\end{tabular}}
\label{tab:ablation}
\vspace{-4mm}
\end{table}

To investigate the effect of different components in VIP, we conduct ablation experiments over PEMS08 by taking one component away each time:
\textbf{VIP wo. extra}: It does not use the dynamic extrapolation in Eq. \eqref{eq:transfer}. Instead, we use an MLP to map the representation $H_{N,T}[b]$ to $H_{N,T}$. \textbf{VIP wo. b}: It removes random regularization of the learnable mask $\hat{b}$. \textbf{VIP wo. p}: It removes random regularization of the learnable mask $\hat{p}$. \textbf{VIP wo. replay}: It removes the prioritized variable-parameter replay.

Table \ref{tab:ablation} compares VIP (wo. or w. pretrain) with all the above variants, each missing a component, over PEMS08 dataset when the deployment ratio is $m/n=10\%$. The degraded result of VIP wo. extra demonstrates the importance to capture the spatial correlations among the locations with the help of adjacency matrix and location embedding. The degraded result of VIP wo. b (VIP wo. p) demonstrates the importance of randomly-initiated sparse learnable node mask $\hat{b}$ (parameter mask $\hat{p}$), which generates different selections for optimization, instead of stable selections that lead to sub-optimal results. Finally VIP wo. replay also performs worse than VIP because it suffers from the catastrophic forgetting problem, where the model cannot remember the previous knowledge during pruning as new knowledge is learned.

\subsection{Diversity Study}\label{sec:diversity study}

\begin{table}[ht]
\centering
\caption{Location selection diversity and forecast accuracy under different regularization methods over all batches from the PEMS08 dataset with 10\% deployment ratio.}

\resizebox{0.8\columnwidth}{!}{
\begin{tabular}{|c|c|c|c|c|}
\hline
\multirow{2}{*}{\textbf{Regularization}} & \multirow{2}{*}{\textbf{Jaccard Distance}} & \multicolumn{3}{c|}{\textbf{VIP w. pretrain}} \\ \cline{3-5}
& & MAE & RMSE & MAPE\\ 
\hline
VIP w. L1  & 2.8\% & 18.59 & 31.33 & 12.63\\ \hline
VIP w. L2  & 2.3\%  &19.19 & 32.59 & 13.31\\ \hline
VIP w. ElasticNet  & 2.4\% & 19.07 & 32.11 & 13.02\\ \hline
VIP wo. Reg  & 1.5\% & 20.25 & 33.90 & 14.53\\ \hline
\end{tabular}}
\label{tab:variance}

\end{table}

In order to justify how L1 norm in Eq.~(\ref{eq:sum loss}) promotes diversity in location selection, we present experiment results in Table~\ref{tab:variance} that compares different regularization methods, including our choice of VIP with L1 regularization (VIP w. L1) and alternative choices of VIP with L2 regularization (VIP w. L2), VIP with Elastic Net regularization \cite{zou2005regularization} (VIP w. ElasticNet), and VIP without any regularization (VIP wo. reg). We focus on two metrics: the forecast accuracy and the Jaccard Distance \cite{real1996probabilistic}, which is defined to measure the diversity of a dataset, in our case, the diversity of the location selections (i.e., variable masks) across all batches in all iterations. Its formula is 
\[
\text{Jaccard Distance} = 1 - \frac{2}{K \cdot B(B-1)} \sum_{k=1}^{K} \sum_{1 \leq t < t' \leq B} \frac{|b_t^k \cap b_{t'}^k|}{|b_t^k \cup b_{t'}^k|}
\]
where $K$ is the total number of iterations to reach the $10\%$ deployment ratio, $B$ is the number of batches for each iteration, $b_t^k$ and $b_{t'}^k$ are variable masks for the $t$-th batch and the $t'$-th batch in the $k$-th iteration. 

The experiment results show that VIP with L1 regularization generates variable masks with the greatest Jaccard Distance, indicating better location selection diversity, and thus achieves the best accuracy when compared to other regularization methods. 

\subsection{Scalability Study}\label{sec:scalability study}
\begin{table}[t]
\centering
\caption{Model parameter size (in MB) under varying numbers of variables.}
\resizebox{0.75\columnwidth}{!}{
\begin{tabular}{|c|c|c|c|c|c|c|c|}
\hline
\textbf{Model} & \textbf{100} & \textbf{500} & \textbf{1,000} & \textbf{5,000} & \textbf{10,000} & \textbf{50,000} & \textbf{100,000} \\
\hline
GinAR & 0.67 & 2.02 & 4.61 & 61.29 & 222.14 & 5108.94 & 20217.44 \\
\hline
VIP   & 0.68 & 1.09 & 1.64 & 7.88 & 20.18  & 298.59  & 1096.59 \\
\hline
\end{tabular}
}
\label{tab:parameters_scalability}
\end{table}

\begin{table}[t]
\centering
\caption{Inference time per forward pass (in milliseconds) under varying numbers of variables.}
\resizebox{0.75\columnwidth}{!}{
\begin{tabular}{|c|c|c|c|c|c|c|c|}
\hline
\textbf{Model} & \textbf{100} & \textbf{500} & \textbf{1,000} & \textbf{5,000} & \textbf{10,000} & \textbf{50,000} & \textbf{100,000} \\
\hline
GinAR & 75.1  & 141.4  & 328.9  & 10086.76 & OOM & OOM & OOM \\
\hline
VIP   & 3.47  & 3.53   & 4.35   & 4.35     & 6.08 & 53.86 & 188.6 \\
\hline
\end{tabular}
}
\label{tab:inference_time_scalability}
\end{table}

We present the results of the scalability studies in Table~\ref{tab:parameters_scalability} and Table~\ref{tab:inference_time_scalability}, comparing the memory footprint and inference latency of our VIP models against the most related GinAR model, respectively. All experiments were conducted on an NVIDIA B200 GPU with 180 GB memory, an Intel(R) Xeon(R) Platinum 8570 CPU, and 200 GB RAM. The batch size is set to 1, $l = l' = 12$, and the deployment ratio is $10\%$ across all scales.

The results show that as the total number $n$ of variables increases from 100 to 100,000, VIP exhibits significantly better scalability in both memory efficiency and inference speed. In Table~\ref{tab:parameters_scalability}, GinAR’s model size grows rapidly from 0.67 MB at 100 variables to 20,217.44 MB at 100,000 variables, while VIP’s model size increases much more slowly—from 0.68 MB to only 1,096.59 MB under the same conditions. In Table~\ref{tab:inference_time_scalability}, VIP's inference latency is 188.6 ms per forward pass even at 100,000 variables, whereas GinAR's inference latency is 10086.76ms at 5,000 variables and it experiences out-of-memory (OOM) errors beyond 10,000 variables due to excessive memory consumption. 

These results clearly demonstrate that VIP scales more favorably than GinAR in both model size and inference time, making it highly suitable for real-time spatio-temporal forecasting in large scale systems.

\subsection{Different Experience Replay Methods}\label{subsec:replay}

To further evaluate the design of our prioritized variable-parameter replay (PVR), we substitute it with two other experience replay designs: random experience replay (RAND-ER) and inverse prioritized variable-parameter replay (IN-PVR). 
%FIFO-ER replays the oldest samples from the buffer.  
RAND-ER randomly replays the samples from the buffer.  IN-PVR gives the high-loss samples a higher probability of being replayed. Specifically,  we flip the priority $P_{S}$ in Eq. \eqref{eq:store sample} to $P_{S} =\mathcal{L}_{main}(\cdot,\cdot)$, and the probability for a sample to be selected is proportional to $P_{S}$.

\begin{table}[t]
\centering
\caption{Different experience replay methods on PEMS08 dataset with a deployment ratio of $m/n=10\%$.}

\resizebox{0.7\columnwidth}{!}{
\begin{tabular}{|c|c|c|c|c|c|c|}
\hline
\multirow{2}{*}{\textbf{Models}} & \multicolumn{3}{c|}{\textbf{VIP wo. pretrain}} & \multicolumn{3}{c|}{\textbf{VIP w. pretrain}}\\ \cline{2-7}
& MAE & RMSE & MAPE & MAE & RMSE & MAPE\\
\hline
PVR & 19.87 & 32.73 & 13.19 & 18.59 & 31.33 & 12.63 \\ \hline
%FIFO-ER & 19.98 & 33.29 & 13.38 & 19.52 & 32.13 & 12.89 \\ \hline
RAND-ER & 20.47 & 33.62 & 13.53 & 19.47 & 31.92 & 13.08\\ \hline
IN-PVR & 21.68 & 34.74 & 18.35 & 20.56 & 33.81 & 15.38\\ \hline
\end{tabular}
}

\label{tab:diff_er}
\end{table}

Table \ref{tab:diff_er} compares PVR and RAND-ER and IN-PVR in terms of average MAE, MAPE and RMSE on dataset PEMS08 with a deployment ratio of $m/n=10\%$. It shows that PVR achieves the best performance because the dynamic change of variable/parameter masks causes model instability, the pruning of variables and parameters can result in forgetting the established patterns, and replaying low-loss samples helps mitigate that. 
%FIFO-ER achieves the second best forecast accuracy because it replays the oldest samples to the current model, which helps maintain a long-term memory of the oldest knowledge. 
That is also the reason that IN-PVR performs worse than RAND-ER because replaying high-loss samples will not do better in restoring model stability during pruning than replaying random samples.

%\textcolor{blue}{add reviewer 1 comment W8}

\begin{figure}
    \centering
    \begin{minipage}{0.35\textwidth}
        \centering
        \includegraphics[width=\textwidth]{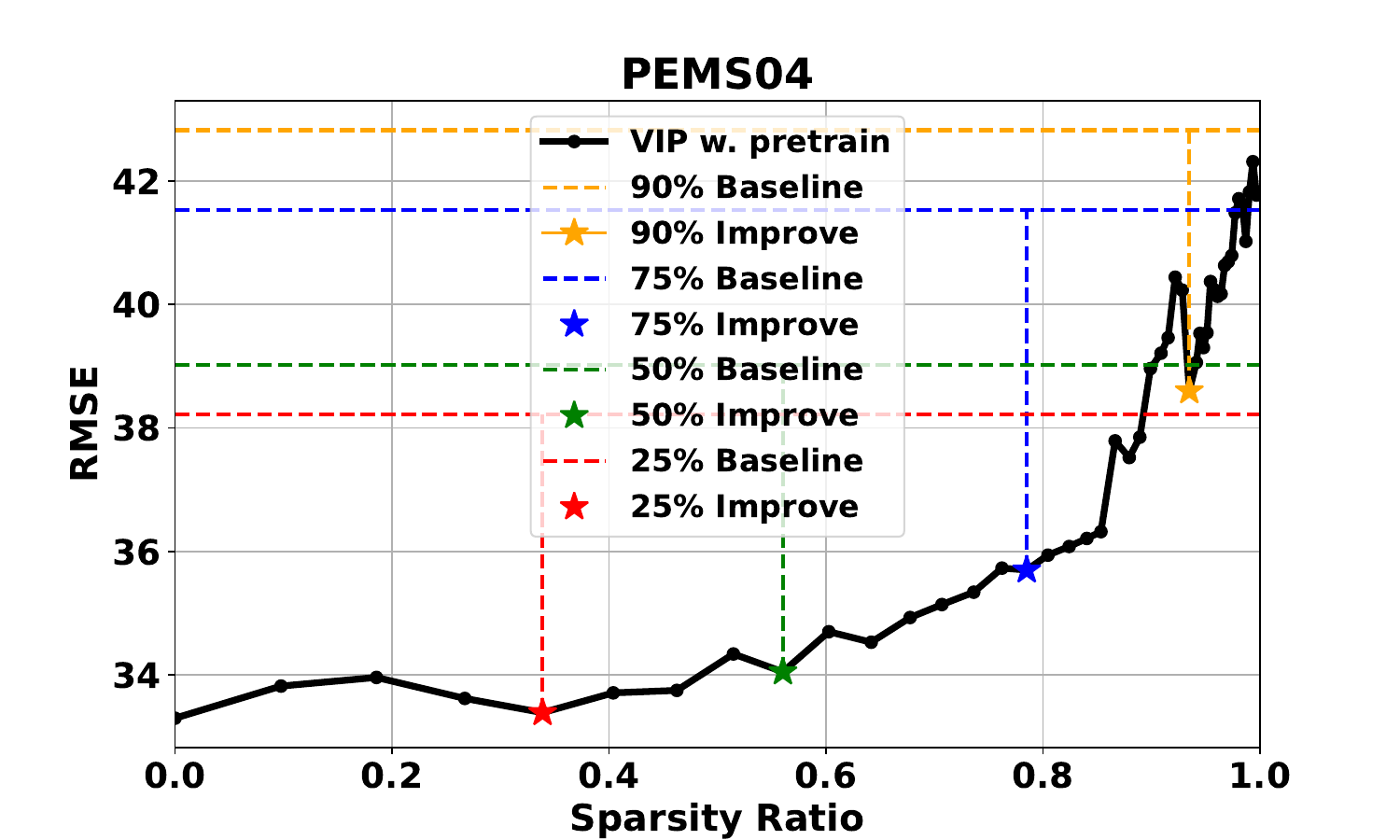}

    \end{minipage}
    \hfill
    \begin{minipage}{0.35\textwidth}
        \centering
        \includegraphics[width=\textwidth]{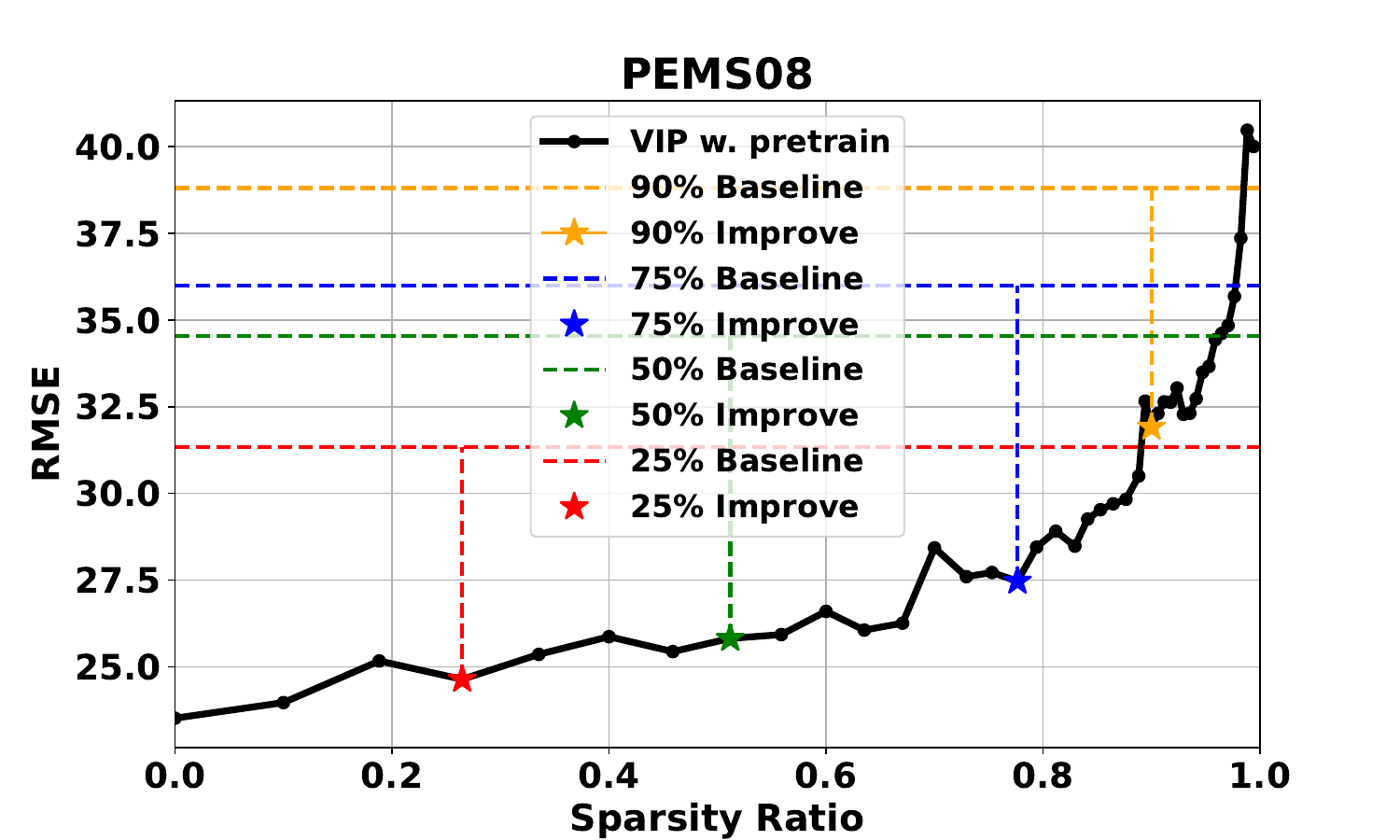}

    \end{minipage}
        \begin{minipage}{0.35\textwidth}
        \centering
        \includegraphics[width=\textwidth]{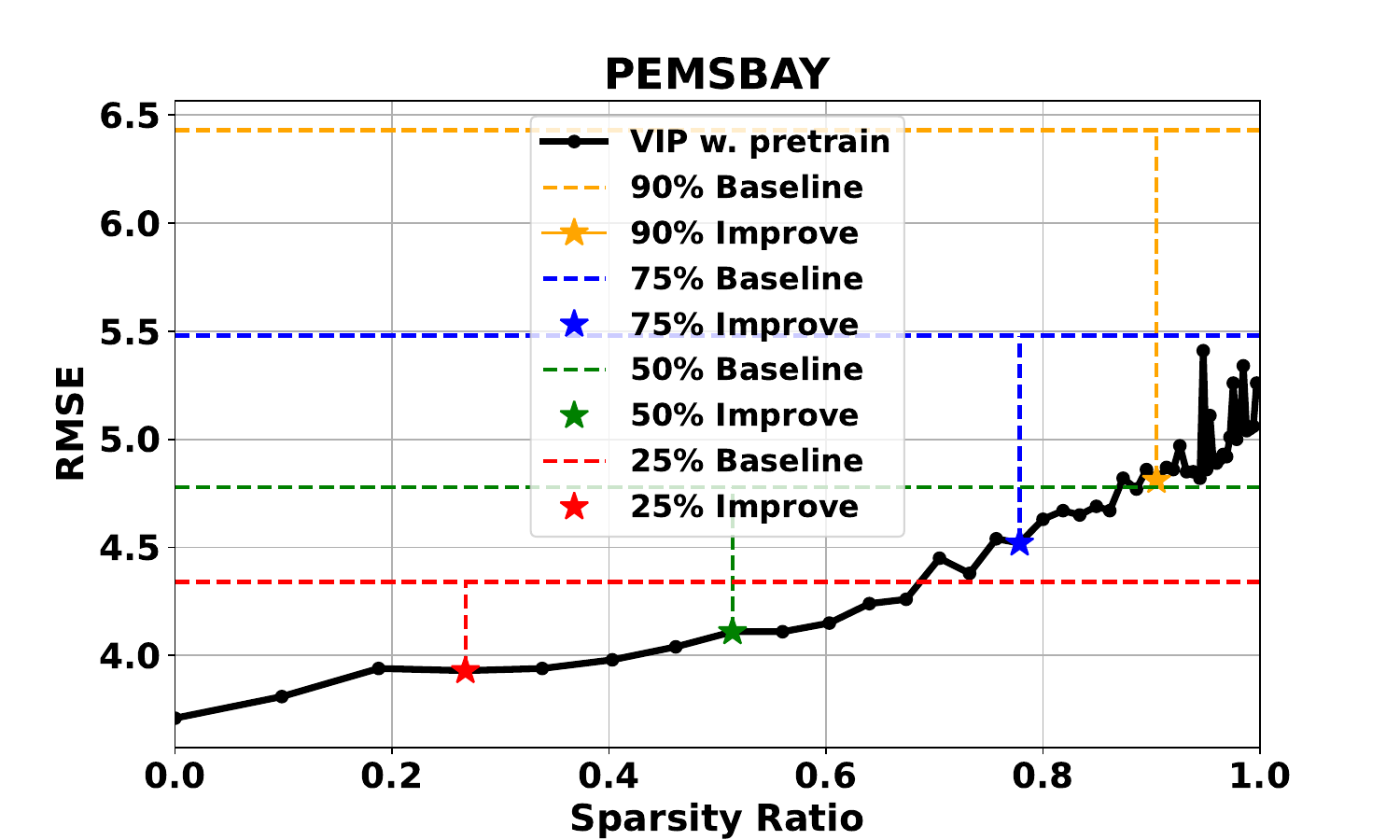}

    \end{minipage}
        \begin{minipage}{0.35\textwidth}
        \centering
        \includegraphics[width=\textwidth]{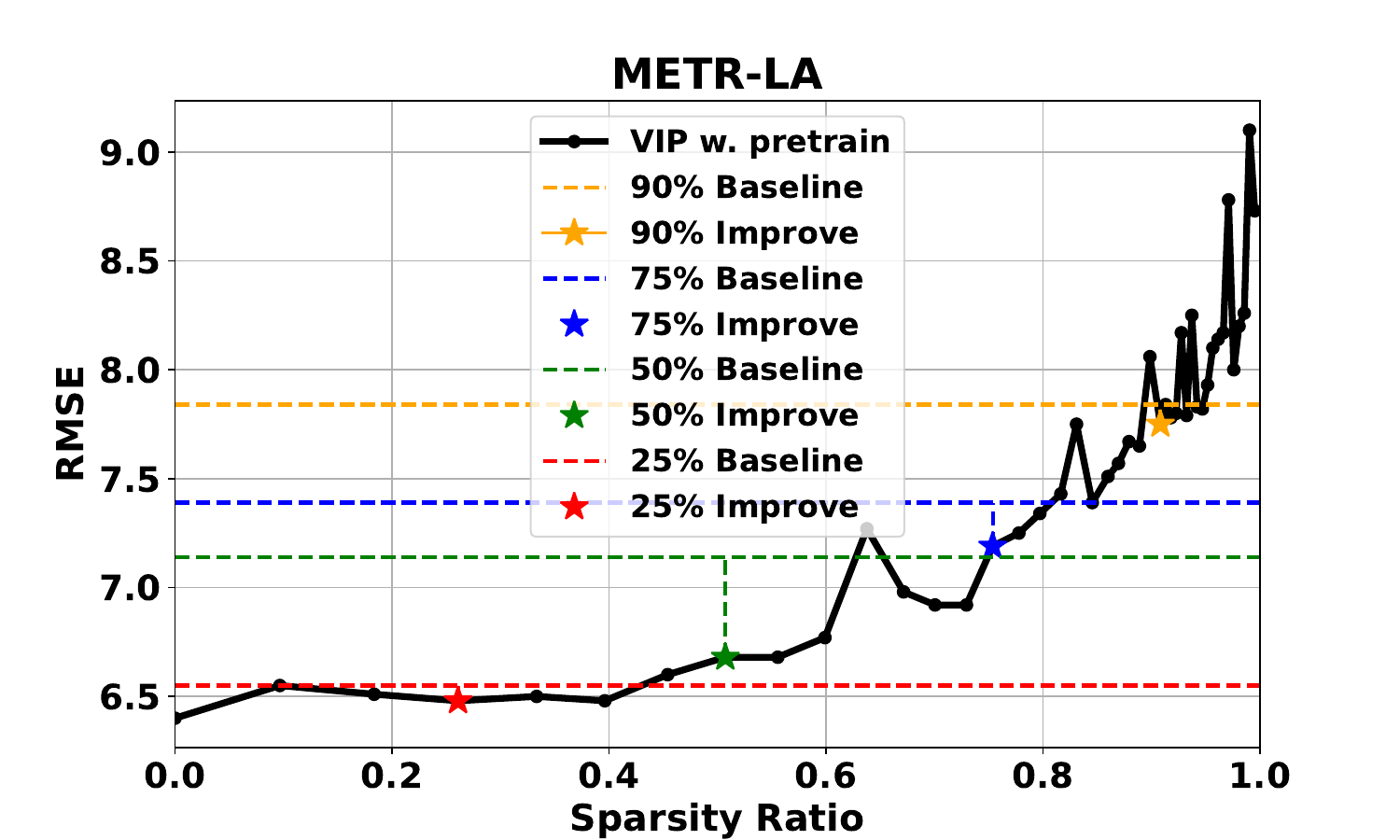}

    \end{minipage}
    \begin{minipage}{0.35\textwidth}
        \centering
        \includegraphics[width=\textwidth]{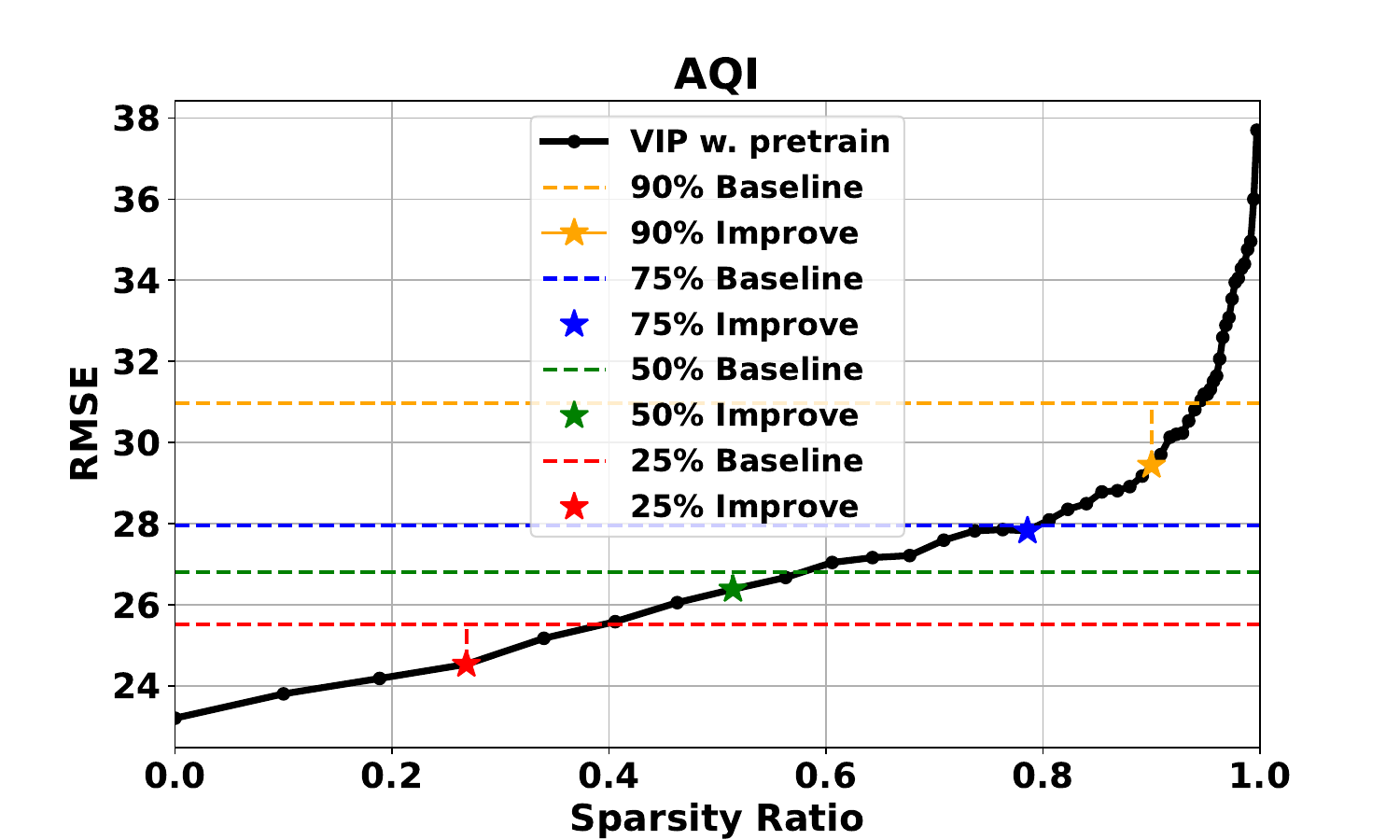}

    \end{minipage}
    \caption{Visualization of node sparsity in different datasets.}
    \label{vis:node sparse}  
\end{figure}

\subsection{RMSE under Different Sparsity Ratios}
Fig. \ref{vis:node sparse} presents the forecast accuracy results (in terms of RMSE) of VIP w. pretrain under different sparsity ratios, with the pruning variable rate $r_b=0.1$ and the pruning parameter rate $r_p=0.05$. 
%When the sparsity ratio is $0\%$, it becomes the traditional STMF, shown in Fig. \ref{fig:fluctuations}(a). 
The baseline for comparison is GinAR, which is the most related work.  
Each plot in the figure is for a different dataset.
The horizontal dashed lines show the RMSE values achieved by the baseline GinAR under sparsity ratios of $25\%$, $50\%$, $75\%$, and $90\%$, respectively. The stars on the curve show the RMSE values achieved by VIP w. pretrain under these ratios, respectively. Each vertical line segment from a star shows the RMSE improvement of VIP over the baseline. The results demonstrate that VIP is more accurate than GinAR under different sparsity ratios over different datasets. Note that a higher sparsity ratio means a lower deployment ratio.

\section{Related Works}\label{sec:related work}
Most existing work focuses on the spatial-temporal forecast problem shown in Figure~\ref{fig:fluctuations}(a), including recurrent neural networks (RNNs) ~\cite{Zhang_Zheng_Qi_2017}, convolutional neural networks (CNNs) or multiple layer perceptrons (MLPs) with graph neural networks (GNNs) \cite{yu2017spatio,li2018diffusion,wu2019graph,li2020spatial,Jiang_Wang_Yong_Jeph_Chen_Kobayashi_Song_Fukushima_Suzumura_2023,yi2023fouriergnn}, neural differential equations (NDEs) \cite{fang2021spatial,liu2023graphbased}, transformers \cite{liu2023spatio,10.1609/aaai.v37i4.25556}, and most recent mixture-of-experts (MOE) \cite{shazeer2017outrageously} based model \cite{lee2024testam}. 

Some recent papers \cite{li2023missing,cini2021filling,chen2023biased,tang2020joint,zhang2023trid} investigate missing-data forecast as shown in Figure~\ref{fig:fluctuations}(b), and they attempt to recover the missing measurements at locations where sensors may sometimes fail. These methods still assume that sensors are deployed at all locations. When data from certain locations remain unavailable for long periods, the existing approaches would introduce incorrect temporal dependencies, leading to suboptimal data recovery performance. Spatial extrapolation methods, such as non-negative matrix factorization methods \cite{lee2000algorithms} and GNN methods \cite{wu2021inductive,zheng2023increase,hu2023graph,appleby2020kriging}, could be used to estimate missing values of some variables based on available values from other variables  within the same time period, but they do not model complex spatio-temporal dynamics for accuracy forecast into future time periods.

A few recent papers study STMF with missing variables, as shown in Figure~\ref{fig:fluctuations}(c), including Frigate \cite{10.1145/3580305.3599357}, STSM \cite{su2024spatial}, and GinAR \cite{yu2024ginar}. In the context of traffic forecast, they assume that only some locations are deployed with permanent sensors. Among them, the most related work is GinAR, which performs much better than Frigate, and unlike STSM it does not require meta-knowledge about location environment (such as nearby malls). However, GinAR assumes the sensor locations (i.e., chosen variables) are given, without handling the issue of optimal senor placement, which is addressed by this paper. 

The problem of sensor deployment has been extensively studied in the domain of wireless sensor networks (WSNs)~\cite{priyadarshi2020deployment}, where the objective is to determine the optimal sensor placement to ensure sensing coverage and communication connectivity in a given spatial area, with each sensor having a sensing radius and a communication radius. Techniques in this domain include force-based models \cite{abidin2017review} (e.g., virtual force algorithm \cite{yu2013node}), grid-based methods \cite{al2013quantifying}, and metaheuristics \cite{gupta2016genetic}. This line of research differs fundamentally from ours: (1) Their goal is to ensure that all interested areas/points/lines are wirelessly covered by the deployed sensors, whereas we do not address the coverage problem and our sensors produce data at the locations of their deployment, without a radius of wireless coverage. (2) Our goal is to use sensed information in the past from a selected subset of locations to predict the information about other locations in the future, whereas their work is not related to that. Therefore, their optimized locations for coverage will not align with our optimal locations for forecast.

Another line of research is the subset selection, which has emerged as a promising direction for improving the efficiency of machine learning by selecting representative data samples for training, thus reducing computational cost while maintaining performance. Notable approaches \cite{pooladzandi2022adaptive,yang2023towards,xia2022moderate} focus on sample-based selection for text or images tasks, however, they barely study the spatio-temporal forecasting problem. %Besides, they do not focus on the reduction of model parameters.  

%Also related is the work on graph lottery tickets (GLT) \cite{yan2024multicoated,chen2021unified,duan2024pre}, which however sparsifies the links of an input graph, not the nodes, whereas the latter is the focus of this paper.

Finally, a component of our VIP framework adopts the experience replay (ER) method \cite{zhang2017deeper}, which stores the historical samples in the buffer and replays the samples to the current model, which is a common technique in the continual learning domain \cite{aljundi2019task} to mitigate catastrophic forgetting \cite{kirkpatrick2017overcoming}. Recently, the prioritized experience replay (PER) \cite{ schaul2015prioritized} follows the ER method and prioritizes the samples to be replayed, achieving excellent performance primarily in the classification task in computer vision and natural language processing. However, its effectiveness in our spatial-temporal forecast context remains unclear.   

In summary, this is the first work on STMF with chosen variables that performs variable selection and parameter reduction to optimize forecast performance in accuracy and efficiency.  

\section{Conclusion}
This paper studies a new problem of spatio-temporal multivariate time series forecast with chosen variables (STCV). Our variable-parameter iterative pruning model (VIP) jointly optimizes variable selection and parameter sparsification via a co-optimized training scheme. By integrating a dynamic extrapolation strategy and a prioritized replay mechanism, VIP maintains stable learning under iterative pruning. Experiments across five real-world datasets validate the effectiveness and efficiency of our proposed model.  

% \newpage

\FloatBarrier
\bibliographystyle{IEEEtran}
\bibliography{reference}

% \input{sections/appendix}
% \input{sections/response}

% \bibliography{response_reference}
% \let\cleardoublepage\clearpage
\end{document}